%% file: faithful_previews_arxiv.tex
\documentclass[10pt,letterpaper]{article}
\usepackage{preprint,times}
\input{math_commands.tex}

\usepackage{amssymb}
\usepackage{url}
\usepackage{graphicx}
\usepackage{wrapfig}
\usepackage{capt-of}
\usepackage{booktabs}
\usepackage{multirow}
\usepackage[hidelinks]{hyperref}
\hypersetup{
  pdftitle={Unlocking Few-Step Diffusion for Faithful Previews},
  pdfauthor={Jing Jia, Sifan Liu, Guanyang Wang},
  pdfsubject={Faithful few-step diffusion previews}
}

\title{Unlocking Few-Step Diffusion for Faithful Previews}
\author{{\large Jing Jia\textsuperscript{1,*}\quad
  Sifan Liu\textsuperscript{2}\quad
  Guanyang Wang\textsuperscript{3,*}}\\[5pt]
  {\small
    \textsuperscript{1}Department of Computer Science, Rutgers University\\
    \textsuperscript{2}Department of Statistical Science, Duke University\\
    \textsuperscript{3}Department of Statistics, Rutgers University\\[4pt]
    \texttt{jing.jia@rutgers.edu}\quad\texttt{sifan.liu@duke.edu}\\
    \texttt{guanyang.wang@rutgers.edu}\\[3pt]
    \textsuperscript{*}Corresponding authors.
  }
}
\date{}

\begin{document}

\maketitle

\noindent
\begin{minipage}{\linewidth}
    \centering
    \includegraphics[width=\linewidth]{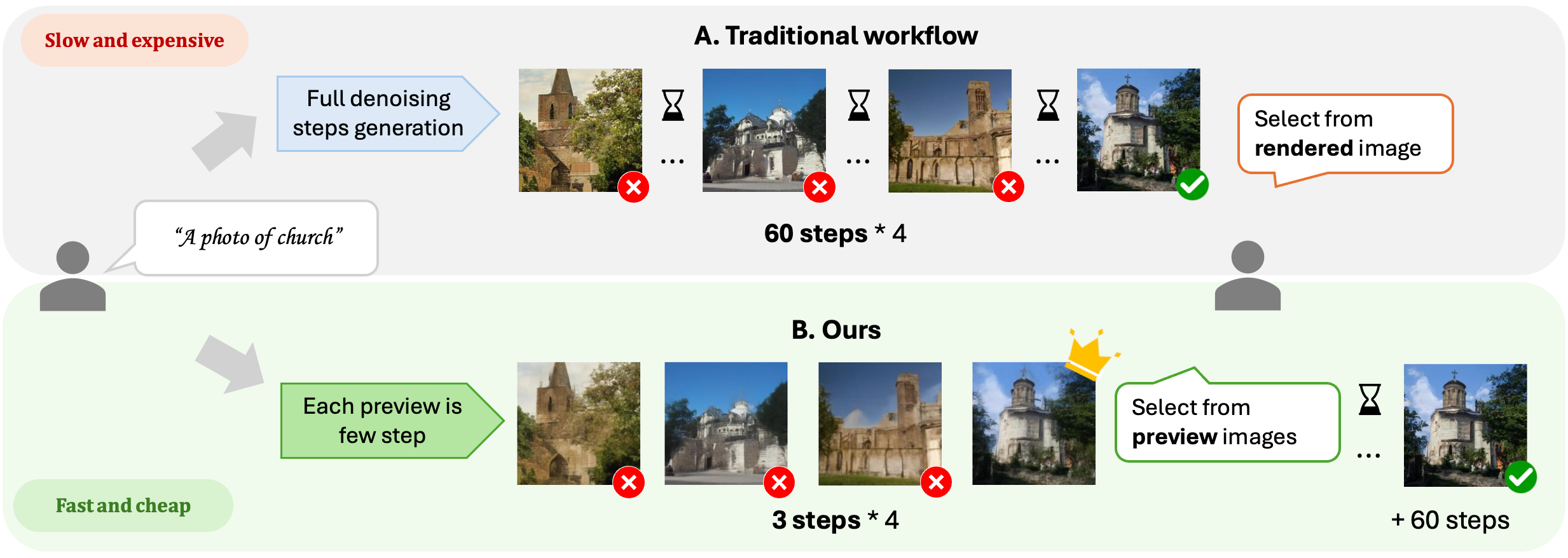}
    \captionof{figure}{\textbf{Traditional vs. preview-then-refinement workflow.}
        Traditional generation requires full denoising steps
        (e.g. 60 steps) before selection, making rejected candidates
        expensive. Our enhancement module enables a faithful
        3-step preview for cheap selection, with full generation
        performed only for selected candidates.
    }
    \label{fig:noise_enhance_workflow}
\end{minipage}

\medskip
\begin{abstract}
Sampling latency compounds in diffusion workflows, where users generate and discard many candidates before keeping one. Surprisingly, the poor outputs of standard few-step samplers do not reflect a lack of reconstruction capacity: by optimizing only the initial noise, frozen 3–4-step samplers can closely reproduce their corresponding full-step outputs. Building on this finding, we learn  corrections to the initial noise and denoising updates using endpoint supervision, improving correspondence with full-step outputs generated from the same noise and prompt. The resulting previews allow users to screen candidates cheaply and reserve full-step generation for promising ones. Input correction also transfers across sampling budgets without retraining. Experiments show substantial improvements in reference fidelity, including 53–78\% lower reconstruction MSE than retrained LD3 on unconditional benchmarks, alongside improved ranking preservation and candidate selection on SD1.5, SDXL, and FLUX.1-dev.
\end{abstract}

\section{Introduction}

Diffusion models generate high-quality images, but users often wait through round after round of sampling before finding the right one. Each image takes tens to hundreds of denoising steps to generate \citep{podell2024sdxl,esser2024scaling,cai2025z,labs2025flux}, and that cost compounds over every round of the search.  Ideally, users could quickly preview candidates and reserve full-step generation for the images they choose to keep, avoiding expensive sampling for candidates they would ultimately discard.

Such previews require \textit{instance-wise correspondence}: the quick preview should be close to the full-step output when using the same initial noise and prompt. However, achieving this correspondence with only a few steps is challenging.
For standard few-step solvers, coarse time discretization can cause
significant deviations from the full-step trajectory. For example,
three- or four-step DDIM sampling often produces blurry images
that differ substantially in appearance and structure from their
full-step counterparts (for example, see the first and second column of
Figure~\ref{fig:oracle_noise}).
Distilled few-step samplers can produce sharper images
\citep{lin2024sdxl,yin2024improved,yin2024one}, but improved sample
quality alone does not ensure that these images match the
corresponding full-step outputs.  For example, DMD2 \citep{yin2024improved} removes the  paired regression loss and improves generation quality through distribution-matching.

Our study was sparked by an observation from diffusion-based inverse problems \citep{wang2024dmplug,jia2026weak}.
A diffusion prior that on its own generates blurry or out-of-domain samples can still recover a corrupted image very well through suitable optimization of the initial noise. While their objective differs from ours, their results hint at an unexplored possibility for few-step samplers:

\textit{The blurry outputs of naive few-step sampling may substantially
underestimate what the frozen sampler can generate. With a suitable
adjustment to the initial noise, few-step samplers may suffice
to closely reproduce the corresponding full-step image.}

We first test this possibility through oracle noise optimization
  with access to full-step reference images. Optimizing only the
  initial noise of frozen samplers reduces mean reconstruction MSE
  by 99.1--99.5\% across four unconditional models with three-step
  sampling, and by 98.6\% for four-step FLUX.1-dev. These results encourage us to learn a shared input corrector that predicts a  residual adjustment to the initial noise. We train it to make the frozen few-step sampler’s output from the corrected noise match the full-step reference generated from the original noise. Across unconditional diffusion benchmarks, our method achieves significantly
  higher fidelity to full-step references than evaluated baselines, such as LD3 \citep{tong2025learning}.
  It also transfers to larger sampling budgets through simple rescaling,
  without retraining.

To extend this approach to large-scale text-to-image models, we explore more flexible trajectory corrections through low-rank adapters. Under controlled experiments, these adapters further improve instance-wise correspondence while keeping the pretrained base weights frozen. 

Returning to our central preview application, corrected few-step samplers let users quickly screen candidates and reserve full-step generation for promising ones. Experiments on SD1.5, SDXL, and FLUX.1-dev show substantial overall gains in ranking preservation and candidate selection over evaluated distilled models, learned preview solvers, and probe-based predictors.

Our contributions are threefold. We reveal a substantial gap between raw sampling quality and reconstruction capacity: even when their raw outputs are severely degraded, frozen 3–4-step samplers can closely reproduce the corresponding full-step generations through oracle optimization of only their initial noise. Building on this finding, we develop  corrections that improve instance-wise correspondence between few-step and full-step generation, with input correction also generalizing across sampling budgets without retraining. The resulting corrected samplers enable fast, faithful previews, allowing users to screen candidates cheaply and reserve full-step generation for promising ones. As an unexpected byproduct, we observe that cross-backbone transfer can also improve perceptual preference scores.

\textbf{Related Works:}
ConsistencySolver \citep{wang2026image} learns integration coefficients through reinforcement learning, whereas we train corrections using direct endpoint supervision. LD3 \citep{tong2025learning} learns sampling schedules with training-only noise perturbations. Diffusion Probe \citep{huang2026diffusion} and Probe-Select \citep{guo2026toward} predict final quality scores from early model features. 

More broadly, other diffusion acceleration methods share our goal of
  computational efficiency but serve complementary purposes. These include
  improved numerical solvers \citep{lu2022dpm} and distillation methods
\citep{salimans2022progressive,song2023consistency,lin2024sdxl,yin2024one,yin2024improved}. These approaches prioritize fast, high-quality generation, which
  alone does not ensure faithful previews. For example, DMD2 removes the loss for instance-wise matching to teacher outputs
  and uses distribution-level objectives to improve sample quality
  \citep{yin2024improved}.

Finally, prior work has shown that refining initial noise can improve generation and  inverse problem solving \citep{eyring2024reno,ahn2026noise,wang2024dmplug,jia2026weak}. We instead target instance-wise fidelity to a full-step teacher for fast
  previews.

 \section{Method}\label{sec:method}

We consider deterministic diffusion samplers that map initial noise $z$, typically drawn from a standard Gaussian, and an optional condition $c$ to an image. Let $F_{\mathrm{ref}}$ denote a fixed full-step reference (or teacher) sampler and $F_{\mathrm{few}}$ a few-step student sampler. Our goal is to correct $F_{\mathrm{few}}$ so that its output closely matches $F_{\mathrm{ref}}(z,c)$ for the same input $(z,c)$. In most of our experiments, the two samplers are based on the same model family (e.g., Stable Diffusion 1.5 or SDXL), although our formulation also allows different model families.

\begin{wrapfigure}[16]{r}{0.50\textwidth}
    \centering
    \includegraphics[width=\linewidth]{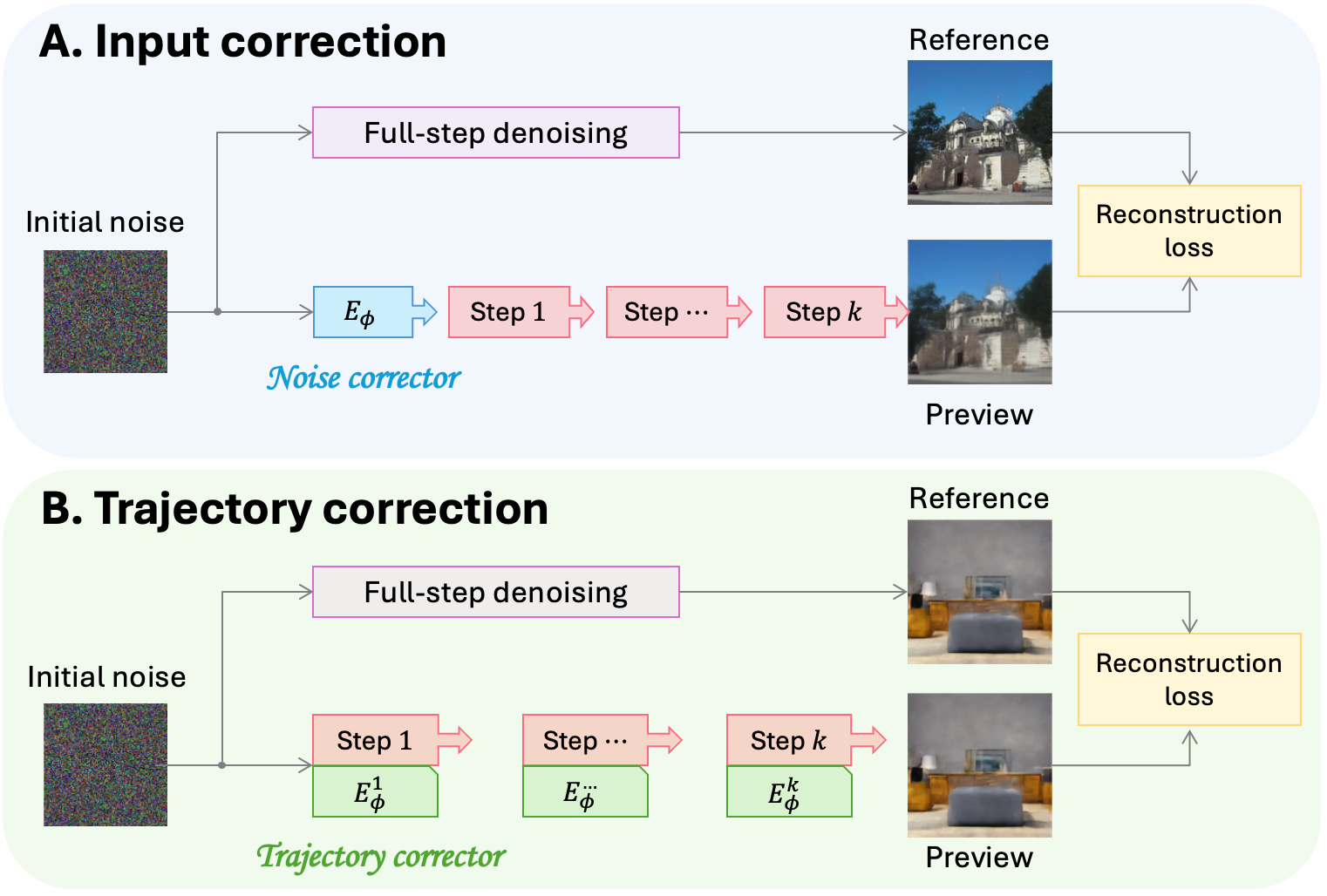}
    \caption{Input correction (A) and trajectory correction (B) for fast and faithful diffusion previews.}
    \label{fig:method}
\end{wrapfigure}
\paragraph{Training objective.}
We denote the corrected few-step map by $\widehat{F}_{\phi}$, where
$\phi$ represents the trainable correction parameters. For each input
$(z,c)$, we use the reference output as the training target and minimize
the error between the two outputs:
\begin{equation}
    \min_{\phi}\;
    \mathbb{E}_{z,c}\!\left[
        \ell\!\left(
            \widehat{F}_{\phi}(z,c), F_{\mathrm{ref}}(z,c)
        \right)
    \right],
\end{equation}
where $\ell$ measures image reconstruction error.
\paragraph{Training procedure.}
We sample inputs $(z,c)$, compute the corresponding reference outputs
$F_{\mathrm{ref}}(z,c)$, and minimize the above objective over $\phi$.
Gradients are propagated through the complete few-step sampler,
while the pretrained backbone remains frozen. Training requires
only input--output pairs from the reference sampler.
Losses, auxiliary terms, and implementation details are provided in
Appendix~\ref{app:correction-details}.

\paragraph{Parameterizing the correction.}
We consider two natural locations for correction: the initial noise
and the intermediate denoising updates.

\emph{Input correction} uses a lightweight network to predict a residual
$\Delta z$ from the initial noise $z$. We then add this residual to $z$
and pass the corrected noise to the fixed few-step sampler:
\[
z \;\longrightarrow\; z+\Delta z
\;\longrightarrow\; F_{\mathrm{few}}(z+\Delta z,c).
\]
Intuitively, this residual adjusts the starting point in a learned
direction that helps the few-step sampler better approximate the
full-step output associated with the original noise.

\emph{Trajectory correction} keeps the original initial noise and
modifies the denoising updates through step-specific low-rank
adapters \citep{hu2022lora}. Each step has its own adapters, which
add trainable low-rank updates to the frozen denoiser weights.
The adapters are trained jointly to improve the final output
of the complete sampling chain.
Figure~\ref{fig:method} illustrates the two
correction strategies for a $k$-step sampler. For trajectory correction, we choose to supervise only the
final outputs. This avoids the need for intermediate
reference states and allows us to treat the reference sampler
as a black box that provides input--output pairs.

 \paragraph{Classifier-free guidance.}
  In text-to-image diffusion, classifier-free guidance (CFG) combines
  conditional and unconditional denoiser predictions using a guidance scale $s$,
  which controls the strength of the text condition. Because changing $s$ also
  changes the generation trajectory, we treat it as an additional input to the
  correction model. We encode $s$ with Fourier features and use a learned
  projection to continuously modulate the correction. This conditioning is
  compatible with both input and trajectory correction.

 \paragraph{Transfer across steps.}
  Input correction learns a sample-dependent direction in the initial-noise space: adjusting the noise along this direction helps a few-step sampler match
  the full-step reference. We find that the learned direction remains useful at
  other sampling budgets without retraining the corrector. For a corrector trained with $k$ sampling steps, we adapt it to a $K$-step
  sampler by scaling its predicted noise displacement by e.g. $\alpha_K=(k/K)^2$. For
  example, suppose a three-step corrector predicts $\Delta z$ such that
  $F_{\mathrm{3-step}}(z+\Delta z)\approx F_{\mathrm{ref}}(z)$. We can use the
  same direction for five-step sampling with input $z+(3/5)^2\Delta z=z+0.36\Delta z$, so that $F_{\mathrm{5-step}}(z+0.36\Delta z)\approx F_{\mathrm{ref}}(z)$. This lets us reuse a trained noise corrector across sampling steps without
  retraining. We demonstrate the effectiveness of this approach in Section~\ref{subsec:noise corr}.

\paragraph{Inference and cheap preview.}
  The corrected few-step sampler is itself a fast generative model.
  Because the corrected sampler preserves instance-wise correspondence with the
  full-step reference, we can naturally use its outputs as faithful previews of
  the reference images generated from the same inputs. Suppose a user wants to find an image they like for a prompt $c$. Instead of running the full-step sampler for every noise in a batch $\{z_i\}$
  before deciding which image to keep, we generate corrected few-step previews
  and use them to select a candidate or discard the batch. If we select a
  candidate, we run the original full-step sampler from its original $(z_i,c)$.  Since the corrector is trained to minimize image reconstruction error, selection is not tied to a particular image metric. Users can apply a metric suited to their task or choose directly from the previews, keeping a human in the loop.

\section{Instance-wise correspondence through few-step samplers}
\subsection{Hidden Reconstruction Capacity of Frozen Few-Step Samplers}

\begin{wrapfigure}{r}{0.40\textwidth}
    \centering
    \includegraphics[width=\linewidth]{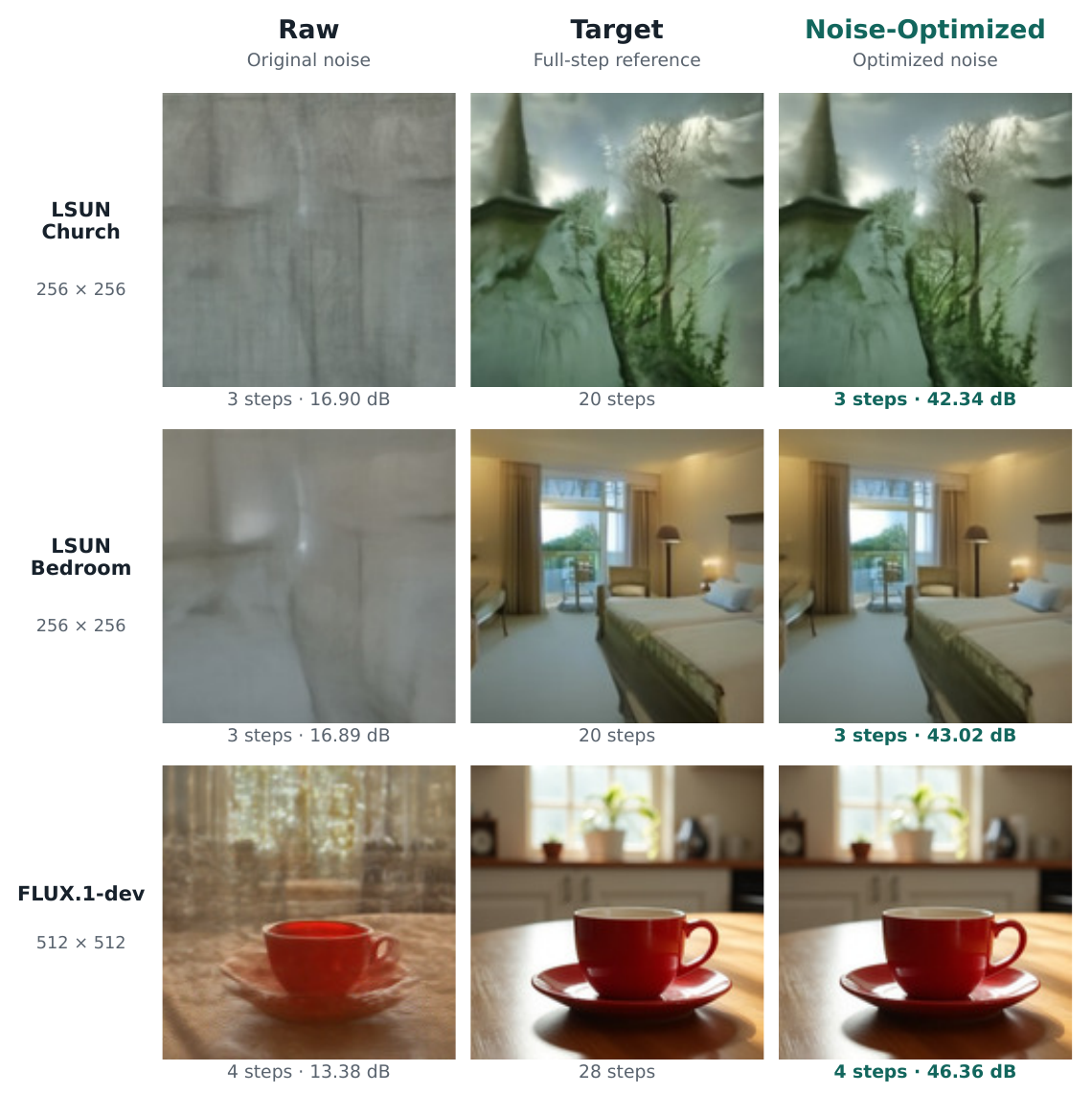}
  \caption{Noise optimization reveals the reconstruction capacity of frozen few-step samplers. Columns show raw outputs, full-step targets, and noise-optimized outputs.}
  \label{fig:oracle_noise}
\end{wrapfigure}

Our approach is inspired by diffusion-based inverse problems
  \citep{wang2024dmplug,jia2026weak}. These methods optimize the
  initial noise of a frozen, few-step diffusion model (e.g., 3-4 step DDIM sampler) to reconstruct an
  image from corrupted observations. Even when a few-step sampler
   produces blurry images (see the first column of Figure \ref{fig:oracle_noise}), optimizing its input noise can
  yield accurate reconstruction. This leads us to ask whether the
  same idea can help generation: can we correct the initial Gaussian
  noise so that a few-step sampler produces an image close to its
  full-step counterpart?

As a proof of concept, we optimize the initial noise of frozen
     three-step DDIM samplers on CIFAR-10, LSUN-Church, CelebA-HQ,
     and LSUN-Bedroom. References use 1,000-step DDIM for CIFAR-10
     and 20-step DDIM for the other three datasets.
     We evaluate 100 seeds per model and perform 200 Adam updates
     per image. Optimization starts from the reference image's
     original noise, with all model parameters frozen and no
     constraint on the optimized noise.

     Across the four unconditional models, noise optimization raises
     mean PSNR from 14.62--17.00\,dB to 37.41--39.98\,dB.
     We additionally evaluate FLUX.1-dev on four fixed prompt--noise
     pairs, using a four-step ODE sampler, 28-step reference targets,
     and 2,560 optimization updates per image, with reference
     guidance 3.5. Mean PSNR increases from 14.54 to 37.49\,dB
     (Table~\ref{tab:oracle_noise}).

     These target-aware experiments demonstrate that frozen few-step
     samplers can closely reproduce full-step outputs through 
     input optimization. They motivate learning an input correction
     that approximates this capability without target access or
     per-image optimization at inference.

\begin{table*}[t]
  \centering
  \small
  \caption{Frozen-model oracle noise optimization. PSNR (dB) is the mean $\pm$ sample SD over $N$ images; few/ref denotes student/teacher steps. }
  \label{tab:oracle_noise}
   \begin{tabular}{lrrrrr}
       \toprule
       Model / data & $N$ & few/ref & Raw
       & Noise-Optimized & Gain (dB) \\
       \midrule
       DDPM / CIFAR-10
       & 100 & 3 / 1000
       & $17.00\pm1.91$ & $37.41\pm1.89$ & +20.41 \\
       DDPM / LSUN-Church
       & 100 & 3 / 20
       & $14.96\pm1.49$ & $38.01\pm3.15$ & +23.05 \\
       DDPM / CelebA-HQ
       & 100 & 3 / 20
       & $14.62\pm1.65$ & $39.14\pm3.89$ & +24.52 \\
       DDPM / LSUN-Bedroom
       & 100 & 3 / 20
       & $15.52\pm1.77$ & $39.98\pm3.09$ & +24.46 \\
       \midrule
       FLUX.1-dev
       & 4 & 4 / 28
       & $14.54\pm1.95$ & $37.49\pm7.69$ & +22.95 \\
       \bottomrule
     \end{tabular}
\end{table*}

\subsection{Input Correction}\label{subsec:noise corr}

  We evaluate input correction on
  CIFAR-10~\citep{krizhevsky2009learning},
  LSUN-Church~\citep{yu2015lsun}, and
  CelebA-HQ~\citep{liu2015faceattributes}.
  For each dataset, the base sampler is 3-step
  DDIM~\citep{song2021denoising} with a frozen pretrained denoising
  network. The reference sampler is 1000-step DDIM for CIFAR-10
  and 20-step DDIM for Church and CelebA-HQ.
  Given an initial noise sample $z$, the learned corrector predicts
  a residual $\Delta z$. We add this residual to $z$, rescale the
  result to preserve the original noise norm, and feed the corrected
  noise into the unchanged 3-step DDIM sampler, as described in
  Section~\ref{sec:method}.
  Training and implementation details are provided in
  Appendix~\ref{app:noise_training}.

  We evaluate all methods on the same 100 held-out noise samples
  per dataset and compare their outputs with the corresponding
  teacher outputs generated from the same initial noises.
  We report MSE and PSNR for pixel fidelity, LPIPS for perceptual
  similarity, and SSIM for structural similarity.
  Metrics are computed at each model's native resolution and
  averaged over individual samples; in particular, the reported
  PSNR is the mean per-image PSNR.

  Our baselines include the uncorrected 3-step DDIM sampler,
  4-step DDIM, 4-step
  DPM-Solver++~\citep{lu2025dpmsolverpp}, and two
  4-step LD3 variants~\citep{tong2025learning}.
  The first LD3 variant applies the same publicly released
  four-step LSUN time-step schedule directly to our  backbone,
  without retraining or schedule remapping. 
  Meanwhile, we also learn a dataset-specific schedule
  on the target backbone using its corresponding teacher.
  Both LD3 variants require four denoiser evaluations.
  Our method instead uses three denoiser evaluations and one
  lighter corrector evaluation, denoted by $3+1$ in
  Table~\ref{tab:correction_results}.

\begin{wrapfigure}{r}{0.45\textwidth}
    \centering
    \includegraphics[width=\linewidth]{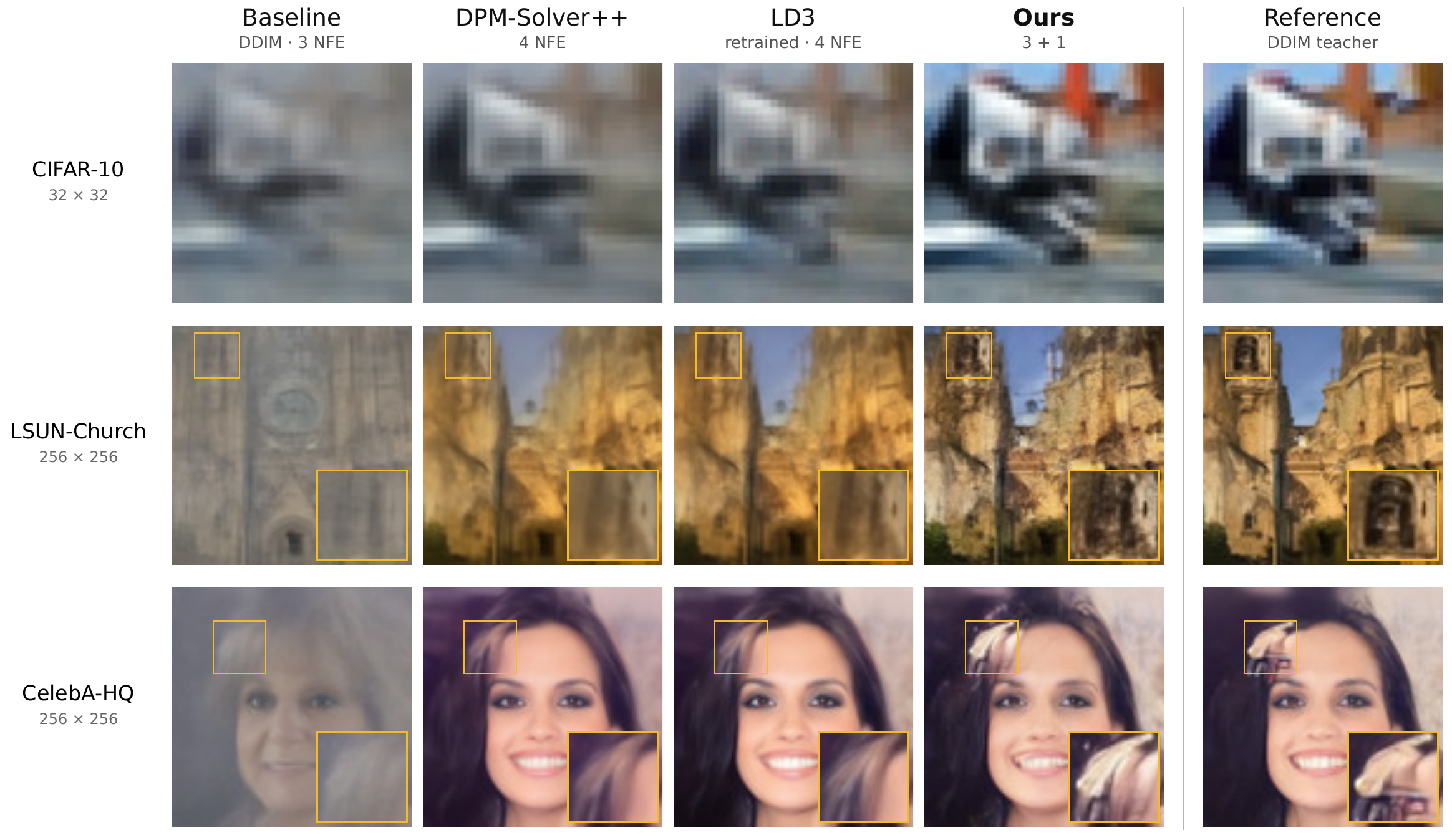}
  \caption{Representative examples of input correction on CIFAR-10, LSUN-Church, and CelebA-HQ.  Columns show uncorrected 3-step DDIM, 4-step DPM-Solver++, 4-step LD3, and our method.}
  \label{fig:noise_correction}
\end{wrapfigure}

  As shown in Table~\ref{tab:correction_results}, input correction
  improves all four teacher-reconstruction metrics over every
  evaluated baseline on all three datasets.
  Relative to uncorrected 3-step DDIM, our method reduces MSE by
  89.7\%, 91.3\%, and 95.8\% on CIFAR-10, Church, and CelebA-HQ,
  respectively.
  Compared with retrained 4-step LD3, the corresponding MSE
  reductions are 78.0\%, 53.2\%, and 71.6\%, accompanied by
  PSNR gains of 3.41--6.75\,dB and LPIPS reductions of
  15.8--44.3\%.
  Thus, under this evaluation protocol, correcting the initial
  noise yields closer agreement with the teacher than the tested
  four-step samplers while retaining the original three-step
  DDIM sampler. Representative examples in Figure~\ref{fig:noise_correction}
  illustrate these differences across all three datasets.
  Our outputs more closely match the teacher's object shape and
  color on CIFAR-10, architectural features on LSUN-Church,
  and facial features and fine details around the hairline on
  CelebA-HQ, whereas the few-step baselines tend to blur or
  alter these details.

 \paragraph{Transfer across steps}
 We investigate whether the correction direction learned for a
  three-step sampler generalizes to larger sampling budgets.
  As more denoising steps become available, we expect a smaller
  correction of the initial noise to be sufficient.
  We therefore reuse the learned correction $\Delta z$ with
  a step-dependent scale $\alpha(K,p)=\left(3/K\right)^p,$ and normalize $z+\alpha(K,p)\Delta z$ to the norm of the
  original noise $z$ before sampling.
  This preserves the full correction at $K=3$ and progressively
  attenuates it as $K$ increases.

  Using the frozen three-step correctors on LSUN-Church and
  CelebA-HQ, we evaluate $K\in\{3,4,6,8\}$ and
  $p\in\{1.7,2.0,2.5\}$ without retraining.
  Each configuration uses the same 200 noises per dataset,
  with 20-step DDIM outputs as paired references.
  As shown in Table~\ref{tab:flexible_steps} in Appendix~\ref{app:extended-transfer-steps} and Figure~\ref{fig:flexible_steps_psnr_hps}, every tested
  configuration improves PSNR over raw DDIM at the same
  denoising step count, while also reducing MSE and LPIPS
  and increasing SSIM, HPS, and ImageReward~\citep{xu2023imagereward}.
  These results demonstrate that the learned correction
  direction remains useful beyond its original sampling budget.

  The exponent provides a simple control over the
  fidelity--quality trade-off.
  At each transferred budget, $p=2.5$ achieves the highest
  PSNR, whereas $p=1.7$ achieves
  the highest HPS.
 A larger exponent favors conservative correction for teacher fidelity, while a smaller exponent favors stronger correction for perceptual quality.

Additional ablations support the use of noise norm normalization,
  sample-specific correction directions, and reduced correction
  magnitudes when transferring to more sampling steps
  (Appendix~\ref{app:subsec:ablation}).
  
\begin{wrapfigure}[17]{r}{0.40\textwidth}
    \centering
    \includegraphics[width=\linewidth]{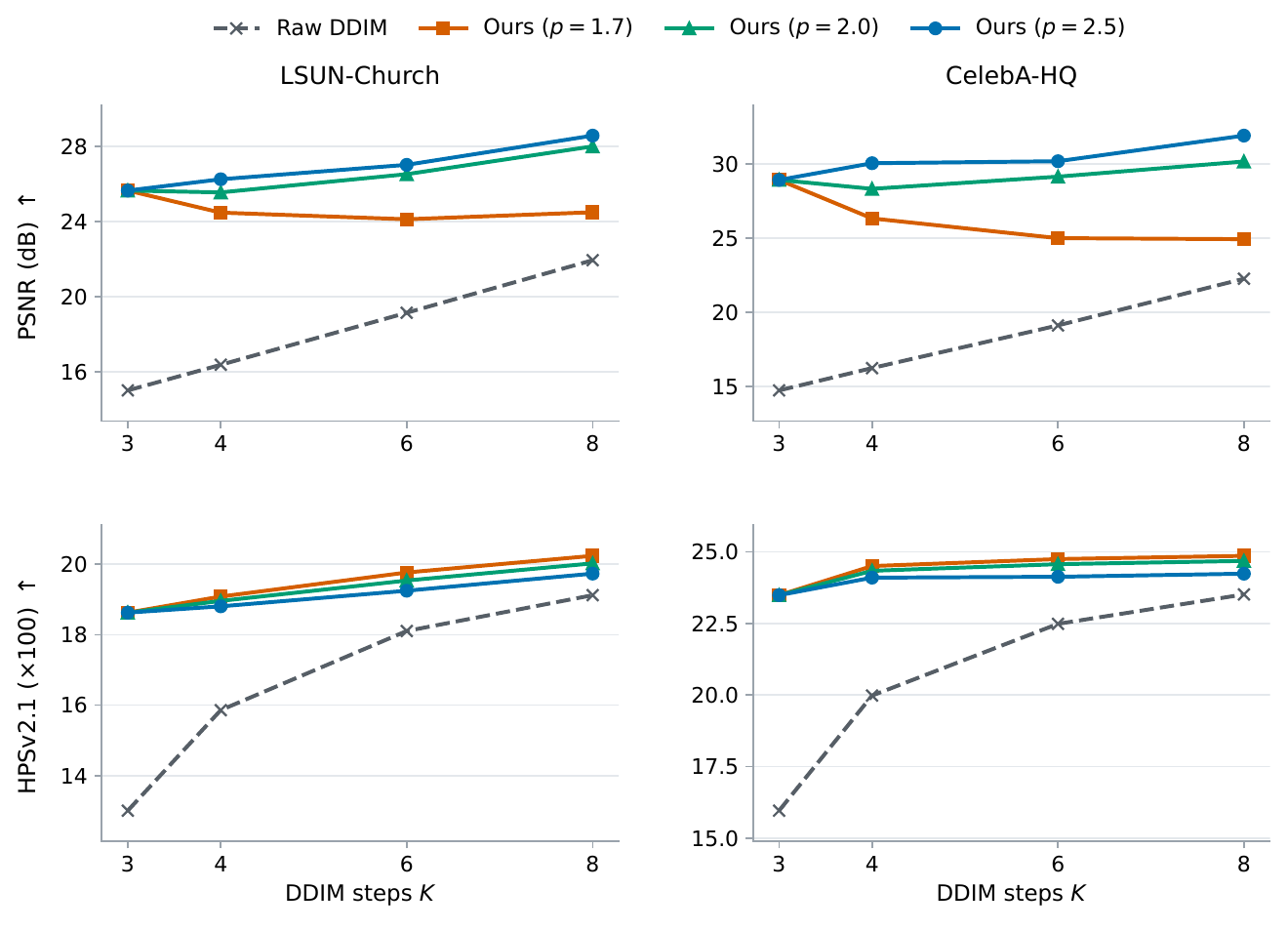}
  \caption{PSNR and HPS on LSUN-Church and CelebA-HQ as the number of
  DDIM steps increases. Dashed lines are raw DDIMs.}
  \label{fig:flexible_steps_psnr_hps}
\end{wrapfigure}

\paragraph{From input to trajectory correction.}
  Input correction improves correspondence while leaving the sampling
  map unchanged. We next ask whether allowing corrections within the
  denoising updates can further improve fidelity. We compare both
  parameterizations on CIFAR-10 and SD1.5 under matched data, endpoint
  objectives, and training-update budgets. Trajectory correction
  improves PSNR over input correction from 26.60 to 29.79\,dB on
  CIFAR-10 and from 20.60 to 23.25\,dB on SD1.5. Both substantially
  outperform uncorrected three-step sampling. These results motivate
  exploring trajectory correction for text-to-image previews; the
  controlled comparison is detailed in
  Appendix~\ref{app:bridge-input-trajectory}.
\subsection{Trajectory Correction}
We next evaluate trajectory correction on SD1.5 and SDXL \citep{Rombach_2022_CVPR, podell2024sdxl}. We train a  trajectory correction on three denoising steps for both models and evaluates on a disjoint set of 100 prompts from COCO val2017 \citep{lin2014microsoft}, each paired with eight fixed noises. We evaluated at different CFG scales comparing with the full-denoising, 60 steps, teacher. We also optimized a 3-step timestamp using LD3 \citep{tong2025learning} for the Stable Diffusion family. For SD1.5, we additionally compare against the ConsistencySolver \citep{wang2026image} pretrained solver. 

Table~\ref{tab:correction_results} reports the SDXL and SD1.5 results at CFG scales $1$, $3$, and $5$. Trajectory correction consistently outperforms all baselines across all four image-level similarity metrics and all evaluated guidance scales. Across SD1.5 and SDXL at CFG scales 1, 3, and 5, averaging per-group improvements against the best reported baseline for each metric, our method reduces LPIPS by 18.3\%, increases SSIM by 18.7\%, and improves PSNR by 2.32 dB. These gains persist as the CFG scale increases; however, stronger classifier-free guidance makes the denoising trajectory increasingly difficult to approximate under aggressive step compression, resulting in larger deviations from the full-denoising teacher. Accordingly, baseline similarity deteriorates at higher CFG scales, while trajectory correction substantially mitigates this degradation. Complete results including higher CFG comparisons are in Appendix~\ref{app:additional_experiments}.
\begin{wrapfigure}{R}{0.35\textwidth}
    \centering
    \includegraphics[width=\linewidth]{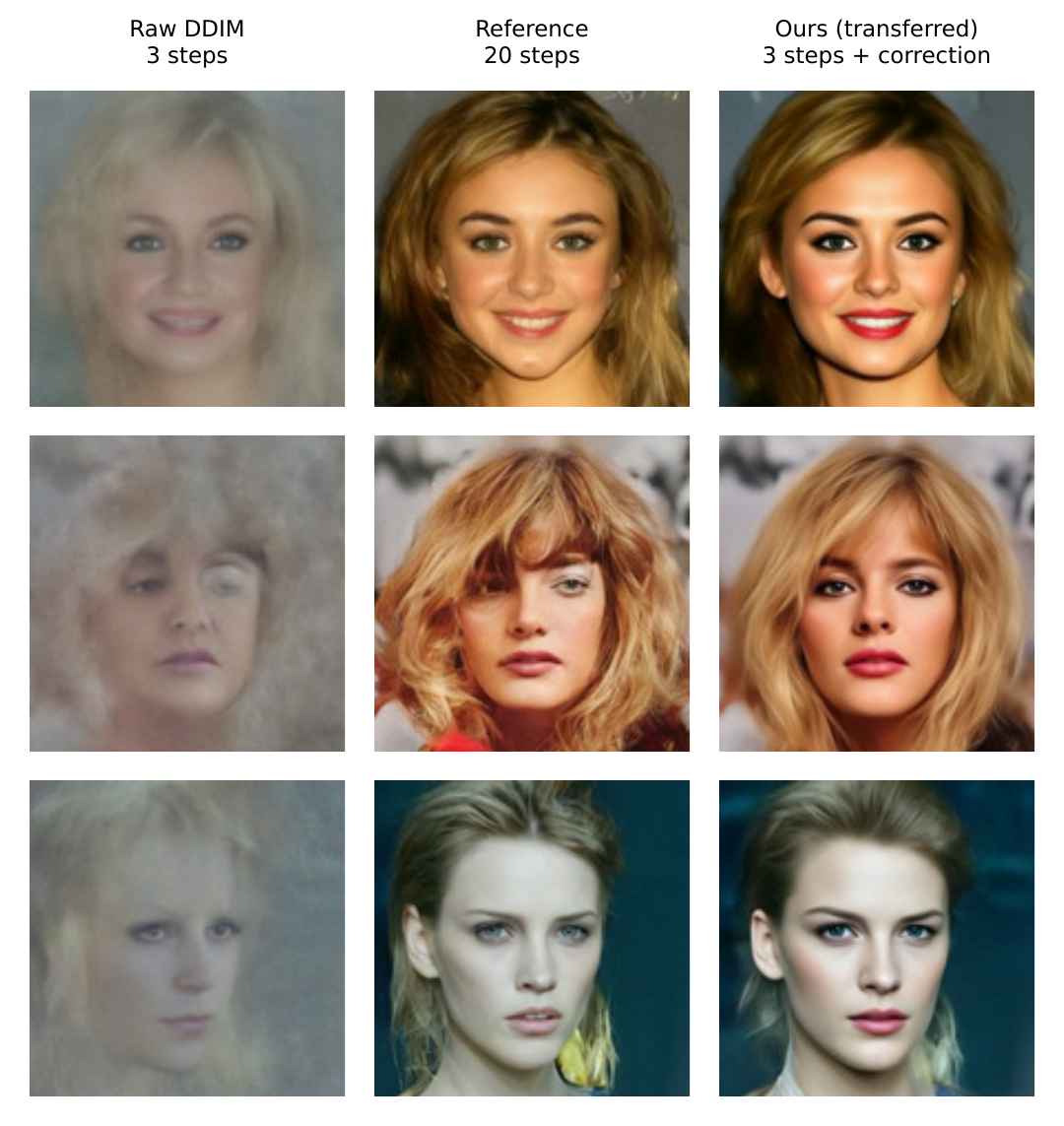}
  \caption{Bedroom-to-CelebA-HQ correction.}
  \label{fig:bedroom_representative}
\end{wrapfigure}

  \begin{table}[t]
  \centering
  \caption{
  Within each dataset/CFG group,
  metrics are reported in the order MSE $\downarrow$, PSNR $\uparrow$,
  LPIPS $\downarrow$, and SSIM $\uparrow$.
  Noise-correction results are averaged over 100 paired test noises
  per dataset.
  For input correction, $3+1$ denotes three denoiser evaluations
  and one lighter corrector evaluation.
  Trajectory correction is evaluated on SDXL and SD1.5.
  Bold denotes the best reported result within each group.}
  \label{tab:correction_results}

  \renewcommand{\arraystretch}{1.08}

    \resizebox{\columnwidth}{!}{\begin{tabular}{@{}l c *{3}{rrrr}@{}}
    \toprule

    \multicolumn{14}{c}{\textit{Input Correction}\qquad
    {\footnotesize
    MSE ($\times10^{-3}$) $\downarrow$\enspace /\enspace
    PSNR $\uparrow$\enspace /\enspace
    LPIPS $\downarrow$\enspace /\enspace
    SSIM $\uparrow$}} \\

    \midrule
    Method & NFE
    & \multicolumn{4}{c}{CIFAR-10}
    & \multicolumn{4}{c}{LSUN-Church}
    & \multicolumn{4}{c}{CelebA-HQ} \\

    \cmidrule(lr){3-6}
    \cmidrule(lr){7-10}
    \cmidrule(lr){11-14}

    3-step DDIM
    & 3
    & 20.79 & 17.281 & 0.355 & 0.575
    & 33.48 & 15.010 & 0.571 & 0.580
    & 34.64 & 14.882 & 0.583 & 0.666 \\

    4-step DDIM
    & 4
    & 15.36 & 18.652 & 0.296 & 0.654
    & 24.11 & 16.447 & 0.482 & 0.632
    & 24.81 & 16.388 & 0.417 & 0.729 \\

    4-step DPM++
    & 4
    & 10.21 & 20.655 & 0.279 & 0.717
    & 7.77 & 21.445 & 0.341 & 0.742
    & 8.14 & 21.563 & 0.301 & 0.812 \\

    4-step LD3 (transferred)
    & 4
    & 24.58 & 17.494 & 0.274 & 0.732
    & 14.16 & 18.708 & 0.500 & 0.712
    & 14.50 & 18.731 & 0.409 & 0.736 \\

    4-step LD3 (retrained)
    & 4
    & 9.75 & 21.045 & 0.248 & 0.764
    & 6.23 & 22.487 & 0.343 & 0.762
    & 5.08 & 23.847 & 0.215 & 0.864 \\

    \textbf{Ours}
    & $3+1$
    & \textbf{2.15} & \textbf{27.790}
    & \textbf{0.138} & \textbf{0.919}
    & \textbf{2.91} & \textbf{25.899}
    & \textbf{0.277} & \textbf{0.856}
    & \textbf{1.44} & \textbf{29.054}
    & \textbf{0.181} & \textbf{0.917} \\

    \bottomrule
    \end{tabular}}

  \par\vspace{0.6em}

    \resizebox{\columnwidth}{!}{\begin{tabular}{@{}l c *{3}{rrrr}@{}}
    \toprule
    
    \multicolumn{14}{c}{\textit{Trajectory Correction} (SDXL)\qquad
    {\footnotesize
    MSE ($\times10^{-3}$) $\downarrow$\enspace /\enspace
    PSNR $\uparrow$\enspace /\enspace
    LPIPS $\downarrow$\enspace /\enspace
    SSIM $\uparrow$}} \\
    
    \midrule
    Method & NFE
    & \multicolumn{4}{c}{CFG = 1}
    & \multicolumn{4}{c}{CFG = 3}
    & \multicolumn{4}{c}{CFG = 5} \\
    
    \cmidrule(lr){3-6}
    \cmidrule(lr){7-10}
    \cmidrule(lr){11-14}
    
    3-step DDIM
    & 3
    & 16.394 & 18.097 & 0.580 & 0.566
    & 26.468 & 15.965 & 0.643 & 0.548
    & 34.866 & 14.733 & 0.676 & 0.518 \\
    
    3-step DPM++
    & 3
    & 10.358 & 20.284 & 0.609 & 0.518
    & 18.285 & 17.800 & 0.572 & 0.628
    & 33.657 & 15.162 & 0.589 & 0.616 \\
    
    3-step LD3 \phantom{ (transferred)}
    & 3
    & 8.038 & 21.416 & 0.632 & 0.564
    & 15.011 & 18.660 & 0.544 & 0.649
    & 28.374 & 15.812 & 0.575 & 0.624 \\
    
    \textbf{Ours}
    & 3
    & \textbf{4.139} & \textbf{24.382}
    & \textbf{0.431} & \textbf{0.727}
    & \textbf{9.010} & \textbf{21.015}
    & \textbf{0.451} & \textbf{0.741}
    & \textbf{13.687} & \textbf{19.154}
    & \textbf{0.477} & \textbf{0.727} \\
    
    \bottomrule
    \end{tabular}}

   \par\vspace{0.6em}

    \resizebox{\columnwidth}{!}{\begin{tabular}{@{}l c *{3}{rrrr}@{}}
    \toprule
    
    \multicolumn{14}{c}{\textit{Trajectory Correction} (SD~1.5)\qquad
    {\footnotesize
    MSE ($\times10^{-3}$) $\downarrow$\enspace /\enspace
    PSNR $\uparrow$\enspace /\enspace
    LPIPS $\downarrow$\enspace /\enspace
    SSIM $\uparrow$}} \\
    
    \midrule
    Method & NFE
    & \multicolumn{4}{c}{CFG = 1}
    & \multicolumn{4}{c}{CFG = 3}
    & \multicolumn{4}{c}{CFG = 5} \\
    
    \cmidrule(lr){3-6}
    \cmidrule(lr){7-10}
    \cmidrule(lr){11-14}
    
    3-step DDIM
    & 3
    & 25.003 & 16.338 & 0.581 & 0.382
    & 31.575 & 15.350 & 0.626 & 0.386
    & 40.469 & 14.234 & 0.661 & 0.384 \\
    
    3-step DPM++
    & 3
    & 17.456 & 18.034 & 0.572 & 0.430
    & 20.485 & 17.292 & 0.546 & 0.509
    & 30.023 & 15.558 & 0.576 & 0.516 \\
    
    3-step ConSolver
    & 3
    & 11.955 & 19.724 & 0.411 & 0.551
    & 18.604 & 17.711 & 0.466 & 0.557
    & 30.309 & 15.482 & 0.535 & 0.527 \\
    
    3-step LD3 \phantom{ (transferred)}
    & 3
    & 16.742 & 18.249 & 0.781 & 0.428
    & 19.521 & 17.569 & 0.635 & 0.510
    & 29.887 & 15.640 & 0.615 & 0.511 \\
    
    \textbf{Ours}
    & 3
    & \textbf{8.585} & \textbf{21.188}
    & \textbf{0.343} & \textbf{0.671}
    & \textbf{13.177} & \textbf{19.377}
    & \textbf{0.399} & \textbf{0.638}
    & \textbf{18.932} & \textbf{17.776}
    & \textbf{0.434} & \textbf{0.615} \\
    
    \bottomrule
    \end{tabular}}

  \end{table}

\subsection{Transfer across backbones}
  \label{sec:unexpected_transfer}
We report an unexpected aesthetic benefit of cross-backbone input correction. Inspired by cross-domain inverse-problem results~\citep{jia2026weak}, we train a corrector through a frozen Bedroom backbone using CelebA teacher targets, then apply it unchanged to CelebA-HQ DDIM3. On 200 test samples, correction improves teacher-reconstruction PSNR from 14.78 to 22.82\,dB. More surprisingly, the outputs exceed both the in-domain corrector and the full-step teacher in HPS, ImageReward, and PickScore \citep{NEURIPS2023_73aacd8b}; for example, HPS increases from the teacher’s 24.88 to 27.34. Figure~\ref{fig:bedroom_representative} illustrates the resulting visual changes, with full results in Appendix~\ref{app:bedroom_transfer} and Figure~\ref{fig:additional-cross-backbone-new-seeds}. We hypothesize that backbone mismatch implicitly regularizes the correction toward perceptually preferred structure. Understanding the mechanism and generality of this effect is of independent interest for future work.

\section{Preservation of Candidate Rankings as Preview}
Because the purpose of a ``cheap preview'' is to identify which candidates merit further computation, image-level reconstruction alone does not fully capture its utility. We therefore evaluate whether few-step previews preserve the relative quality ordering of candidate generations induced by their full-step teacher counterparts. For each prompt, eight noises define a common candidate pool. We independently score the few-step previews and the corresponding full-step teacher images, and measure how well the resulting candidate rankings agree. We consider automated scoring functions including PickScore, ImageReward v1.0, Aesthetic Score, CLIP, and HPS v2.1~\citep{huang2026diffusion, guo2026toward, hessel2021clipscore, wu2023human}. For the Stable Diffusion family, we use GPT-5.6-luna as an agent-based proxy for human preference in place of HPS v2.1, with details in Appendix~\ref{app:agent_evaluation}.

We study the Stable Diffusion family and FLUX.1-dev separately. For SD1.5 and SDXL, we compare against LD3 \citep{tong2025learning}; for SD1.5, we additionally compare against ConsistencySolver \citep{wang2026image}. For FLUX.1-dev, we compare against the probe-based methods DiffusionProbe~\citep{huang2026diffusion} and ProbeSelect~\citep{guo2026toward}, ConsistencySolver, and distilled FLUX variants (TDD-FLUX and FLUX.1-schnell).

\begin{table*}[t]
\centering
\caption{Candidate-ranking preservation on SDXL and SD1.5 across CFG scales. We compare DDIM, DPM++, LD3, and our method, in addition ConSolver for SD1.5. For PickScore, ImageReward, and agent, we report the selected teacher score $\uparrow$, regret $\downarrow$, and Spearman rank correlation $\uparrow$.}
\label{tab:sdxl_ranking}
\resizebox{\textwidth}{!}{
\begin{tabular}{llccc|ccc|ccc|ccc}
\toprule

& & \multicolumn{12}{c}{
SDXL \qquad
  Selected $\uparrow$\enspace /\enspace
  Regret $\downarrow$\enspace /\enspace
  Spearman $\uparrow$
} \\
  
\cmidrule(lr){3-14}

Metric & Method
& \multicolumn{3}{c}{CFG = 1}
& \multicolumn{3}{c}{CFG = 3}
& \multicolumn{3}{c}{CFG = 5}
& \multicolumn{3}{c}{CFG = 7} \\
\midrule

\multirow{4}{*}{\textit{PickScore}}
& 3-step DDIM
& 19.642 & 1.377 & 0.016
& 21.724 & 0.915 & 0.009
& 22.162 & 0.855 & -0.003
& 22.312 & 0.814 & -0.011 \\

& 3-step DPM++
& 20.385 & 0.634 & 0.403
& 22.152 & 0.486 & 0.351
& 22.435 & 0.582 & 0.268
& 22.423 & 0.703 & 0.162 \\

& 3-step LD3
& 20.516 & 0.505 & 0.476
& 22.108 & 0.529 & 0.368
& 22.515 & 0.506 & 0.280
& 22.597 & 0.531 & 0.223 \\

& \textbf{Ours}
& \textbf{20.621} & \textbf{0.397} & \textbf{0.593}
& \textbf{22.215} & \textbf{0.424} & \textbf{0.463}
& \textbf{22.658} & \textbf{0.359} & \textbf{0.466}
& \textbf{22.741} & \textbf{0.386} & \textbf{0.347} \\

\midrule

\multirow{4}{*}{\textit{ImageReward}}
& 3-step DDIM
& -1.200 & 1.242 & -0.030
& 0.274 & 0.701 & -0.022
& 0.550 & 0.598 & 0.003
& 0.646 & 0.529 & 0.021 \\

& 3-step DPM++
& -0.547 & 0.588 & 0.390
& 0.474 & 0.501 & 0.225
& 0.719 & 0.429 & 0.194
& 0.739 & 0.435 & 0.207 \\

& 3-step LD3
& \textbf{-0.332} & \textbf{0.376} & 0.522
& 0.616 & 0.358 & 0.356
& 0.745 & 0.402 & 0.265
& 0.849 & 0.326 & 0.254 \\

& \textbf{Ours}
& -0.363 & 0.404 & \textbf{0.559}
& \textbf{0.728} & \textbf{0.247} & \textbf{0.500}
& \textbf{0.927} & \textbf{0.221} & \textbf{0.390}
& \textbf{0.913} & \textbf{0.261} & \textbf{0.352} \\

\midrule
\multirow{4}{*}{\textit{Agent}}
& 3-step DDIM
& 57.670 & 28.340 & 0.113
& 73.890 & 16.240 & 0.004
& 76.680 & 14.300 & 0.035
& 76.730 & 14.520 & 0.025 \\

& 3-step DPM++
& 70.630 & 15.380 & 0.355
& 81.640 & 8.490 & 0.280
& 81.460 & 9.520 & 0.178
& 81.200 & 10.050 & 0.096 \\

& 3-step LD3
& 72.770 & 13.240 & 0.392
& 81.950 & 8.180 & 0.273
& 80.520 & 10.460 & 0.243
& 80.370 & 10.880 & 0.174 \\

& \textbf{Ours}
& \textbf{77.410} & \textbf{8.600} & \textbf{0.518}
& \textbf{82.700} & \textbf{7.430} & \textbf{0.342}
& \textbf{83.480} & \textbf{7.500} & \textbf{0.290}
& \textbf{82.420} & \textbf{8.830} & \textbf{0.228} \\

\bottomrule
\end{tabular}
}
\par\vspace{0.8em}

\resizebox{\textwidth}{!}{\begin{tabular}{llccc|ccc|ccc|ccc}
\toprule

& & \multicolumn{12}{c}{SD~1.5 \qquad
Selected $\uparrow$\enspace /\enspace
Regret $\downarrow$\enspace /\enspace
Spearman $\uparrow$
} \\
\cmidrule(lr){3-14}

Metric & Method
& \multicolumn{3}{c}{CFG = 1}
& \multicolumn{3}{c}{CFG = 3}
& \multicolumn{3}{c}{CFG = 5}
& \multicolumn{3}{c}{CFG = 7} \\
\midrule

\multirow{5}{*}{\textit{PickScore}}
& 3-step DDIM
& 19.573 & 0.848 & 0.128
& 20.830 & 0.882 & 0.007
& 21.421 & 0.614 & 0.075
& 21.412 & 0.764 & -0.001 \\

& 3-step DPM++
& 19.857 & 0.564 & 0.369
& 21.111 & 0.600 & 0.324
& 21.404 & 0.632 & 0.247
& 21.583 & 0.593 & 0.119 \\

& ConSolver
& \textbf{20.025} & \textbf{0.396} & 0.503
& 21.133 & 0.578 & 0.355
& 21.354 & 0.682 & 0.208
& 21.542 & 0.634 & 0.143 \\

& 3-step LD3
& 19.882 & 0.539 & 0.443
& 21.157 & 0.554 & 0.361
& 21.511 & 0.525 & 0.262
& 21.558 & 0.618 & 0.226 \\

& \textbf{Ours}
& 19.975 & 0.446 & \textbf{0.507}
& \textbf{21.222} & \textbf{0.488} & \textbf{0.449}
& \textbf{21.587} & \textbf{0.449} & \textbf{0.412}
& \textbf{21.680} & \textbf{0.496} & \textbf{0.334} \\

\midrule

\multirow{5}{*}{\textit{ImageReward}}
& 3-step DDIM
& -1.015 & 1.029 & 0.058
& -0.078 & 0.811 & 0.060
& -0.002 & 0.870 & -0.009
& 0.128 & 0.799 & 0.019 \\

& 3-step DPM++
& -0.573 & 0.587 & 0.408
& 0.173 & 0.560 & 0.330
& 0.228 & 0.641 & 0.192
& 0.365 & 0.562 & 0.171 \\

& ConSolver
& \textbf{-0.487} & \textbf{0.501} & \textbf{0.491}
& 0.295 & 0.439 & 0.339
& 0.263 & 0.605 & 0.211
& 0.367 & 0.561 & 0.136 \\

& 3-step LD3
& -0.579 & 0.593 & 0.421
& 0.179 & 0.554 & 0.362
& 0.281 & 0.588 & 0.276
& 0.345 & 0.582 & 0.215 \\

& \textbf{Ours}
& -0.569 & 0.583 & 0.482
& \textbf{0.381} & \textbf{0.352} & \textbf{0.444}
& \textbf{0.391} & \textbf{0.478} & \textbf{0.355}
& \textbf{0.524} & \textbf{0.403} & \textbf{0.278} \\

\midrule
\multirow{5}{*}{\textit{Agent}}
& 3-step DDIM
& 57.770 & 23.810 & 0.169
& 67.160 & 19.880 & 0.069
& 71.070 & 15.990 & 0.047
& 70.330 & 17.390 & 0.075 \\

& 3-step DPM++
& 66.030 & 15.550 & 0.387
& 73.130 & 13.910 & 0.319
& 73.660 & 13.400 & 0.279
& 72.510 & 15.210 & 0.120 \\

& ConSolver
& \textbf{69.810} & \textbf{11.770} & \textbf{0.473}
& 74.160 & 12.880 & 0.340
& 73.660 & 13.400 & 0.221
& 73.280 & 14.440 & 0.154 \\

& 3-step LD3
& 63.210 & 18.370 & 0.423
& 74.320 & 12.720 & \textbf{0.373}
& 74.720 & 12.340 & 0.279
& 73.410 & 14.310 & 0.241 \\

& \textbf{Ours}
& 68.680 & 12.900 & 0.450
& \textbf{76.620} & \textbf{10.420} & 0.357
& \textbf{75.750} & \textbf{11.310} & \textbf{0.294}
& \textbf{76.360} & \textbf{11.360} & \textbf{0.283} \\

\bottomrule
\end{tabular}}
\end{table*}

\subsection{Stable Diffusion Family} 

For each scoring function, we report three statistics. Regret, defined as $s_{\text{oracle}}-s_{\text{selected}}$, which is the difference between the highest teacher score among the eight candidates and the teacher score of the selected candidate. Selected Score is the mean teacher score of the candidate selected by maximizing its preview score, directly measuring the quality achieved by the selection procedure. Lastly, Spearman correlation measures the agreement between the candidate ranking induced by the few-step previews and that of the corresponding full-step teacher generations. 

As reported in Table \ref{tab:sdxl_ranking}, our method consistently preserves teacher rankings across guidance scales and model families. Averaged over the three scoring metrics and four CFG scales in Table 3, Spearman correlation increases from 0.298 for ConSolver to 0.387 for our method on SD1.5, and from 0.319 for LD3 to 0.421 on SDXL. The agent-based evaluation shows a similar trend, with our previews exhibiting higher agreement with the teacher-sample rankings under a rubric that jointly considers prompt adherence, composition, visual quality, and artifacts. Visual examples are also provided in Figure~\ref{fig:sd15_cfg1_group_000051}, \ref{fig:sdxl_cfg1_group_000039}, and \ref{fig:sdxl_cfg5_group_000019}, which illustrate how the previews align with the corresponding teacher samples. Overall, our previews remain informative for candidate selection even under stronger guidance.

\subsection{Image Selection on FLUX.1-dev}
  \label{sec:flux-seed-selection}
 We extend our evaluation to the larger-scale FLUX.1-dev model to test whether cheap previews can identify promising candidates before
  full-step generation. We train a FLUX.1-dev student with trajectory LoRA
  adapters to generate four-step previews, using a 28-step reference sampler
  to provide training targets. Each preview is trained to match the reference
  image generated from the same prompt and initial noise. At inference, we
  score eight previews, select one candidate, and generate its final image with
  the original 28-step sampler.
  
   We compare against two probe-based early quality predictors, a learned
  preview solver, and a distilled model. Diffusion
  Probe~\citep{huang2026diffusion} applies frozen released weights to attention
  features extracted after six denoising steps, using a 20-channel input
  mapping selected in an earlier ImageReward comparison.
  Probe-Select~\citep{guo2026toward} scores step-six transformer features with
  a FLUX-specific head trained on 35,332 image--score pairs. Both probes
  select one seed after six steps and continue its trajectory for the
  remaining 22 steps. We train a four-step FLUX.1-dev adaptation of
  ConSolver~\citep{wang2026image}, optimizing its time-conditioned solver
  coefficients with PPO and a DINOv2 consistency reward against the 28-step
  reference, while keeping the generator frozen. We also evaluate
  TDD-FLUX using the released TDD-ADV LoRA weights with four Euler steps without additional training.

  Our method, ConSolver, and TDD-FLUX perform metric-specific selection:
  each scoring metric independently selects a candidate from the eight
  previews, and we evaluate the corresponding 28-step reference image.  Probe training is itself metric-specific:
  predicting another metric requires a correspondingly supervised prediction
  head, whereas the same image previews can be rescored with different metrics
  without retraining.

  We use 512 COCO val2017 prompts and eight fixed initial noise samples per
  prompt, with all methods selecting from the same candidate pool. Final
  candidates are generated with FLUX.1-dev at $512\times512$ resolution using
  28 Euler steps and guidance embedding 3.5. We also report the expected
  score under uniform random selection and a full best-of-eight oracle
  that completes all eight images before selecting the best one for each
  metric. Further experimental details are given in
  Appendix~\ref{app:flux-details}.

  \begin{table*}[t]
    \centering
    \small
    \setlength{\tabcolsep}{4pt}
    \caption{FLUX.1-dev selection. HPS v2.1 and CLIP-L/14 cosine: $\times100$.
    Bold: best non-oracle mean. NFE counts screening and final-generation
    denoiser calls, excluding decoding/scoring.}
    \label{tab:flux-seed-selection}
    \vspace{\baselineskip}
    \begin{tabular}{lrrrrrr}
      \hline
      Method
      & ImageReward $\uparrow$
      & PickScore $\uparrow$
      & HPS v2.1 $\uparrow$
      & CLIP $\uparrow$
      & AES $\uparrow$
      & NFE $\downarrow$ \\
      \hline
      Random
      & 0.9483 & 22.9300 & 29.6458 & 25.7252 & 5.6658 & 28 \\
      Diffusion Probe
      & 0.9731 & 22.9311 & 29.7302 & 25.7060 & 5.6749 & 70 \\
      Probe-Select
      & 0.9442 & 22.9432 & 29.6611 & 25.7982 & 5.6709 & 70 \\
      ConSolver
      & 1.0085 & 22.9818 & 29.8980 & 25.7976 & 5.6857 & 60 \\
      TDD-FLUX
      & 1.0331 & 23.0183 & 29.9616 & 25.9744 & 5.7166 & 60 \\
      Ours
      & \textbf{1.1417} & \textbf{23.1869} & \textbf{30.3263}
      & \textbf{26.4693} & \textbf{5.7625} & 60 \\
      \hline
      Full best-of-8
      & 1.3640 & 23.5260 & 31.4015 & 27.8651 & 5.9418 & 224 \\
      \hline
    \end{tabular}
  \end{table*}

  Table~\ref{tab:flux-seed-selection} shows that our preview-based selection
  achieves the highest non-oracle mean on all five metrics.
  Our method reaches 1.1417 ImageReward, compared with 1.0331 for
  TDD-FLUX and 1.0085 for ConSolver. The paired ImageReward gain over
  TDD-FLUX is 0.1086, with a 95\% prompt-bootstrap confidence interval
  of $[0.0654,0.1509]$. Thus, our corrected previews provide more effective
  candidate selection than the evaluated distilled model at the same
  sampling budget. All three four-step preview workflows require
  $8\times4+28=60$ NFE, compared with $8\times6+22=70$ for the probes
  and $8\times28=224$ for full best-of-eight.

Additional experiments show that our previews improve candidate selection over FLUX.1-schnell (another distilled model) and an equal-budget raw-sampling  baseline, while generalizing to a higher resolution and PartiPrompts without retraining
  (Appendix~\ref{app:flux-additional}).

\section{Discussion, limitations, and future directions}
Our results reveal underused reconstruction capacity in frozen few-step samplers, with learned input corrections remaining useful across sampling budgets and backbones. We apply these findings to fast previews for candidate selection, improving ranking preservation and selected-image scores over the evaluated baselines. By allowing users to screen previews and reserve full-step generation for selected candidates, this preview--select--render workflow can potentially reduce the time needed to obtain a satisfactory image.

Several limitations and open questions remain. Although our learned correctors substantially improve fidelity, they do not yet match the reconstruction accuracy achieved by target-aware oracle noise optimization, leaving considerable room to explore more effective correction methods. The generality of our empirical findings also warrants further evaluation on larger-scale diffusion models. Moreover, while transfer across sampling budgets and backbones shows promising results, broader experiments and mechanistic investigation are needed to understand when such transfer succeeds and which properties of the learned corrections enable it.

Future work could extend corrected few-step samplers to image editing and inverse problems, use lightweight corrections to help smaller models approximate larger teachers for resource-constrained deployment, and identify backbone–target mismatches that reliably improve fidelity or perceptual quality.

\section*{Acknowledgments}
We thank Haochen Ji and Peng Zhang for helpful discussions.

\bibliography{references}
\bibliographystyle{references}

\newpage
\appendix
\section{Correction Objectives and Method Details}
\label{app:correction-details}
\subsection{Input Correction Details}
 The corrector predicts a residual $\Delta z=C_\phi(z)$.
  For the unconditional experiments, we normalize the corrected input
  to preserve the original noise norm:
  \begin{equation}
      \widetilde z
      = \|z\|_2
        \frac{z+\alpha\Delta z}{\|z+\alpha\Delta z\|_2},
      \qquad
      \widehat x=F_{\mathrm{few}}(\widetilde z).
  \end{equation}
  Normalization is performed independently for each sample, with
  a small denominator floor for numerical stability.
  We use $\alpha=1$ during training.
  The reconstruction target is $F_{\mathrm{ref}}(z)$, associated
  with the original noise. Gradients propagate through the complete
  few-step sampler, while its parameters remain frozen.

\subsection{Trajectory Correction Details}
 Each denoising step uses an independent set of LoRA adapters.
  For a frozen weight matrix $W$, step $i$ uses
  \begin{equation}
      W_i=W+\frac{a}{r}B_iA_i,
  \end{equation}
  where $r$ is the adapter rank and $a$ is its scaling parameter.
  All adapters are trained jointly through the complete student
  rollout. Intermediate student states are not detached or replaced
  with reference states.

  Training supervises both the final output and the terminal latent:
  \begin{equation}
      \mathcal L
      =\mathbb E\left[
        0.1\log\!\left(\mathrm{MSE}_{\mathrm{RGB}}+10^{-6}\right)
        +0.05\,\mathrm{MSE}_{\mathrm{latent}}
      \right].
  \end{equation}
  The logarithm is applied per sample before batch averaging.
  RGB errors use continuous outputs of the frozen VAE decoder.
  This objective requires reference endpoint latents and images,
  but no intermediate reference states.
\subsection{Classifier-Free Guidance Conditioning}
 For Stable Diffusion, we encode the guidance scale $s$ using
  Fourier features and a learned projection, whose output modulates
  the correction. Training targets use the corresponding reference
  guidance scale. FLUX instead uses its native guidance embedding:
  the reference uses 3.5 and the trained student uses 1.0,
  without a separate unconditional branch.
\subsection{Transfer Across Sampling Steps}\label{app:subsec:transfer-steps}
For a corrector trained with $k$ steps, we use
  $\alpha=(k/K)^p$ at inference with $K$ steps.
  Scaling is applied before norm normalization.
  The corrector remains frozen.
  Our experiments use $k=3$, $K\in\{3,4,6,8\}$,
  and $p\in\{1.7,2.0,2.5\}$.
  This transfer procedure applies to input correction. 
\section{Experimental Setup and Implementation Details}\label{app:exp setup}
\paragraph{Unconditional diffusion backbones.}  For CIFAR-10, we use pretrained unconditional DDPM backbones~\citep{NEURIPS2020_4c5bcfec}
  with EMA weights. For LSUN-Church, CelebA-HQ,
  and LSUN-Bedroom, we use
  \texttt{google/ddpm-ema-church-256},
  \texttt{google/ddpm-ema-celebahq-256}, and
  \texttt{google/ddpm-ema-bedroom-256}, respectively,
  all publicly available on Hugging Face.

\paragraph{Conditional diffusion backbones.}
  For text-to-image generation, we use Stable Diffusion 1.5, SDXL,
  and FLUX.1-dev. Their pretrained checkpoints are available on
  Hugging Face as
  \texttt{stable-diffusion-v1-5/stable-diffusion-v1-5},
  \texttt{stabilityai/stable-diffusion-xl-base-1.0}, and
  \texttt{black-forest-labs/FLUX.1-dev}, respectively.

\subsection{Oracle Noise Optimization}\label{app:subsec:oracle}
For unconditional models, we optimize the original Gaussian
     noise through a frozen three-step DDIM sampler using 200 Adam
     updates per image and 100 seeds per dataset. Reference targets
     use 1,000-step DDIM for CIFAR-10 and 20-step DDIM for
     LSUN-Church, CelebA-HQ, and LSUN-Bedroom.

  The FLUX oracle uses four prompt--noise pairs, four-step students, and 2,560 updates.
\subsection{Input Correction Training}
\label{app:noise_training}
For LSUN-Church, we use a residual U-Net corrector, whereas for CelebA-HQ we additionally include attention blocks. Both work well. All students use DDIM timesteps \([666,333,0]\). Reference budgets are 1,000 steps for CIFAR-10 and 20 steps for Church and CelebA-HQ.

\subsection{Trajectory Correction Training}\label{app:subsec:trajectory}

 For SDXL and SD1.5, we use a three-step sampler with timesteps $[666,333,0]$ and
  independent rank-64 LoRA adapters at each step.
  The CFG scale is encoded using Fourier features, followed by a
  learned linear projection with 64 outputs to produce continuous
  rank-wise gates:
  \[
  g(s)=1+\tanh\!\left(W\gamma(s)+b\right).
  \]
  These gates attenuate or amplify the contribution of each LoRA rank.
  Training runs for 8,192 optimization steps on 4,000 prompts,
  each evaluated at five CFG scales.
  Testing uses 100 held-out prompts with eight seeds per prompt
  at each CFG scale.

The FLUX student uses four independent rank-64 adapter sets and 36,864 training pairs.
  We use AdamW with constant learning rate $10^{-4}$,
  weight decay $10^{-4}$, global batch size 64,
  gradient clipping at norm 5, and EMA decay 0.99.
  Training completes 1,548 updates; the EMA checkpoint at update 1,280 is selected using reconstruction PSNR on 128 validation pairs.

\subsection{Agent-Based Visual-Quality Evaluation}
\label{app:agent_evaluation}

We use a multimodal language model as an agent-based proxy for human visual
preference. The evaluator considers prompt adherence, composition, perceptual
quality, and visible artifacts. This evaluation complements conventional
automated metrics and is not intended to replace a human study.

\paragraph{Evaluation protocol.}
For each text prompt and generation method, we collect the eight candidate
images produced from eight fixed initial noises. All eight images are presented
simultaneously to the evaluator in a single multimodal request. Before each
request, we randomly permute their presentation order to reduce positional
bias.

After shuffling, each image is assigned an opaque identifier that allows us to
map the returned evaluation back to its original noise index. The identifiers
do not reveal the generation method, sampling budget, guidance scale, original
position, or noise seed. The evaluator receives only the original text prompt,
the eight candidate images, and the evaluation instruction given below.

We use \texttt{GPT-5.6-luna} as the evaluator. A fresh conversation is created
for every eight-image group, preventing information from being carried across
prompts, methods, or candidate groups. The evaluator first assigns an
independent score between 0 and 100 to every image and then produces a complete
ranking of the eight candidates. Rank 1 denotes the best candidate.

Preview candidates and their corresponding full-step reference candidates are
evaluated as separate eight-image groups. The image order is independently
randomized for each group. The results are subsequently matched through their
underlying noise indices, rather than through their presentation positions.
Thus, the evaluator is not shown explicit preview--reference pairs and is not
asked to perform direct image matching.

\paragraph{Evaluator instruction.}
We use the following instruction for all candidate groups without
method-specific modification:

\begin{quote}

You are a strict, repeatable visual-quality evaluator for text-to-image
research. Score each candidate independently against the supplied prompt, then
rank the candidates within this prompt. Do not reward a candidate for being
more colorful, brighter, or more photorealistic unless that helps satisfy the
prompt. Do not compare candidates across different prompts.

Scoring rubric (0-100):
  90-100: excellent prompt adherence, composition, visual quality, and no
          meaningful artifacts;
  75-89:  strong image with only minor omissions or defects;
  50-74:  recognizable but with noticeable prompt, composition, or quality
          problems;
  25-49:  major omissions, poor composition, or substantial artifacts;
   0-24:  largely unrelated, unusable, or severely corrupted.

Use the full range when justified. Grade each image independently first; then
rank by grade and visual judgment. Rank 1 is best. Break close ties by prompt
adherence, then by visible image quality. Return only the requested JSON.

\end{quote}

The required JSON response contains the opaque identifier, numerical score,
and rank of every candidate. We verify that all eight submitted identifiers
appear exactly once, that every score lies between 0 and 100, and that the
returned ranks form a permutation of $\{1,\ldots,8\}$.

\subsection{FLUX.1-dev Image Selection}
\label{app:flux-details}
We use 512 COCO prompts and eight noises per prompt.
  Final images use 28 Euler steps at $512\times512$ resolution.

For Diffusion Probe, we use the released \texttt{base.pth}
  checkpoint without retraining, with a fixed mapping from
  FLUX attention features to its required 20-channel input. Using the released checkpoint with our feature mapping, Diffusion
  Probe achieves an ImageReward of 0.9731 on our $512\times512$
  FLUX.1-dev benchmark, compared with 0.9483 for random selection.
  Exploratory retraining under our protocol did not yield a
  statistically significant improvement over this baseline:
  the highest score across four runs was 0.9785, a paired gain
  of 0.0054 (95\% confidence interval $[-0.0392, 0.0506]$).
  The original study~\citep{huang2026diffusion} also reports a modest
  ImageReward gain of 0.04, from 1.02 to 1.06, for FLUX.1-dev
  seed selection under its $1024\times1024$ setting with
  25 denoising steps and ten candidates per prompt.

  For Probe-Select, the public implementation we used was
  configured for SD3.5, and we did not find a directly usable
  FLUX checkpoint.
  We therefore train its released prediction-head architecture
  from scratch on FLUX features using 35,332 image--score pairs
  with ImageReward supervision, while keeping the FLUX backbone
  frozen.
  Thus, our comparison uses a frozen public checkpoint for
  Diffusion Probe and a locally trained FLUX adaptation of
  Probe-Select.

  For ConSolver~\citep{wang2026image}, we adapt the released flow-matching
  implementation, originally targeting FLUX Kontext, to four-step
  FLUX.1-dev text-to-image generation. We train  its 
  time-conditioned coefficient MLP, keeping the FLUX backbone, VAE, and
  DINOv2 reward model frozen. We train for 1,000 PPO updates, each using 16 stochastic rollouts of
  one prompt--noise pair. 
  The reward is $50(1+\cos(f_{\mathrm{preview}},f_{\mathrm{reference}}))$,
  where $f$ denotes the DINOv2-base CLS embedding and the reference is
  the 28-step Euler image generated from the same prompt and noise.
  We use AdamW with learning rate $10^{-4}$ and weight decay $10^{-3}$,
  four PPO epochs per update, clipping threshold 0.2, entropy coefficient
  0.01, and policy softmax temperature 0.01.

\section{Additional Experiments}
\label{app:additional_experiments}
\subsection{Additional SD Family Results}
Here we report the trajectory correction results for SD 1.5 and SDXL under high CFG values (7 and 9). Across both models, our method consistently achieves the best performance at both CFG settings, with substantially lower MSE and LPIPS and higher SSIM. The improvements are particularly pronounced at CFG = 9, demonstrating that trajectory correction remains effective and robust even under the increased difficulty induced by stronger classifier-free guidance.

\resizebox{\columnwidth}{!}{\begin{tabular}{lc*{2}{cccc}}
\toprule

\multicolumn{10}{c}{\textit{Trajectory Correction} (SDXL)\qquad
{\footnotesize
MSE ($\times10^{-3}$) $\downarrow$\enspace /\enspace
PSNR $\uparrow$\enspace /\enspace
LPIPS $\downarrow$\enspace /\enspace
SSIM $\uparrow$}} \\

\midrule
Method & NFE
& \multicolumn{4}{c}{CFG = 7}
& \multicolumn{4}{c}{CFG = 9} \\

\cmidrule(lr){3-6}
\cmidrule(lr){7-10}

3-step DDIM
& 3
& 42.634 & 13.846 & 0.775 & 0.485
& 50.090 & 13.141 & 0.790 & 0.456 \\

3-step DPM++
& 3
& 51.495 & 13.301 & 0.630 & 0.574
& 69.432 & 11.980 & 0.671 & 0.530 \\

3-step LD3
& 3
& 44.219 & 13.856 & 0.628 & 0.583
& 59.412 & 12.543 & 0.675 & 0.549 \\

\textbf{Ours}
& 3
& \textbf{18.263} & \textbf{17.873}
& \textbf{0.505} & \textbf{0.709}
& \textbf{22.513} & \textbf{16.922}
& \textbf{0.531} & \textbf{0.692} \\

\bottomrule
\end{tabular}}

\resizebox{\columnwidth}{!}{\begin{tabular}{lc*{2}{cccc}}
\toprule

\multicolumn{10}{c}{\textit{Trajectory Correction} (SD~1.5)\qquad
{\footnotesize
MSE ($\times10^{-3}$) $\downarrow$\enspace /\enspace
PSNR $\uparrow$\enspace /\enspace
LPIPS $\downarrow$\enspace /\enspace
SSIM $\uparrow$}} \\

\midrule
Method & NFE
& \multicolumn{4}{c}{CFG = 7}
& \multicolumn{4}{c}{CFG = 9} \\

\cmidrule(lr){3-6}
\cmidrule(lr){7-10}

3-step DDIM
& 3
& 49.510 & 13.324 & 0.689 & 0.377
& 58.440 & 12.570 & 0.710 & 0.366 \\

3-step DPM++
& 3
& 42.248 & 14.037 & 0.610 & 0.504
& 55.536 & 12.803 & 0.644 & 0.487 \\

3-step ConSolver
& 3
& 43.357 & 13.879 & 0.592 & 0.496
& 56.324 & 12.700 & 0.637 & 0.470 \\

3-step LD3
& 3
& 44.489 & 13.852 & 0.632 & 0.487
& 59.591 & 12.514 & 0.666 & 0.460 \\

\textbf{Ours}
& 3
& \textbf{24.290} & \textbf{16.636}
& \textbf{0.460} & \textbf{0.596}
& \textbf{29.337} & \textbf{15.758}
& \textbf{0.483} & \textbf{0.579} \\

\bottomrule
\end{tabular}}

\subsection{Extended Transfer Across Sampling Steps}
\label{app:extended-transfer-steps}
We evaluate whether input correctors trained with three-step DDIM
  remain effective at larger sampling budgets without retraining.
  For $K\in\{3,4,6,8\}$, we scale the predicted displacement by
  $\alpha=(3/K)^p$, with $p\in\{1.7,2.0,2.5\}$, and normalize
  the corrected input to the original noise norm.
  All configurations use the same 200 noise samples per dataset,
  with paired 20-step DDIM outputs as references.
  As shown in Table~\ref{tab:flexible_steps}, every tested correction
  improves all six metrics over raw DDIM at the same step count.
  At transferred budgets, $p=2.5$ yields the lowest MSE and LPIPS
  and the highest PSNR, whereas $p=1.7$ consistently achieves
  the highest HPS.
  These results show that the learned correction remains useful
  beyond its training budget, with its strength controlling
  a trade-off between reference fidelity and preference scores.

\begin{table*}[htbp!]
  \centering
  \caption{Transfer across steps on 200 samples per dataset.
  HPS: HPSv2.1 $\times100$; IR: ImageReward.
  Best values per dataset, $K$, and metric are bold.}
  \label{tab:flexible_steps}

  \setlength{\tabcolsep}{2.5pt}
  \renewcommand{\arraystretch}{1.08}

  \resizebox{\textwidth}{!}{\begin{tabular}{c l *{2}{rrrrrr}}
  \toprule
  \multicolumn{14}{c}{\textit{Metric order:}
  MSE $\downarrow$ \quad PSNR $\uparrow$ \quad
  LPIPS $\downarrow$ \quad SSIM $\uparrow$ \quad
  HPS $\uparrow$ \quad IR $\uparrow$} \\
  \midrule
  $K$ & Method
  & \multicolumn{6}{c}{LSUN-Church}
  & \multicolumn{6}{c}{CelebA-HQ} \\
  \cmidrule(lr){3-8}\cmidrule(lr){9-14}

  3 & Raw DDIM
  & 0.0336 & 15.024 & 0.571 & 0.577 & 13.010 & -1.985
  & 0.0355 & 14.738 & 0.592 & 0.658 & 15.958 & -1.250 \\

  & Ours (all $p$)
  & \textbf{0.0031} & \textbf{25.661} & \textbf{0.287} & \textbf{0.850} &
  \textbf{18.624} & \textbf{-0.976}
  & \textbf{0.0015} & \textbf{28.909} & \textbf{0.183} & \textbf{0.915} &
  \textbf{23.482} & \textbf{-0.096} \\

  \addlinespace[3pt]
  4 & Raw DDIM
  & 0.0245 & 16.386 & 0.489 & 0.626 & 15.859 & -1.330
  & 0.0254 & 16.242 & 0.425 & 0.720 & 19.976 & -0.488 \\

  & Ours ($p=1.7$)
  & 0.0039 & 24.486 & 0.244 & 0.864 & \textbf{19.081} & -0.770
  & 0.0026 & 26.318 & 0.160 & 0.904 & \textbf{24.499} & \textbf{-0.135} \\

  & Ours ($p=2.0$)
  & 0.0031 & 25.562 & 0.235 & \textbf{0.872} & 18.957 & -0.782
  & 0.0017 & 28.309 & 0.146 & 0.922 & 24.333 & -0.142 \\

  & Ours ($p=2.5$)
  & \textbf{0.0027} & \textbf{26.262} & \textbf{0.233} & 0.869 & 18.803 &
  \textbf{-0.767}
  & \textbf{0.0012} & \textbf{30.035} & \textbf{0.137} & \textbf{0.930} & 24.094
  & -0.142 \\

  \addlinespace[3pt]
  6 & Raw DDIM
  & 0.0131 & 19.157 & 0.348 & 0.738 & 18.103 & -0.776
  & 0.0133 & 19.114 & 0.282 & 0.806 & 22.481 & -0.309 \\

  & Ours ($p=1.7$)
  & 0.0044 & 24.132 & 0.210 & 0.874 & \textbf{19.762} & \textbf{-0.502}
  & 0.0035 & 24.988 & 0.156 & 0.897 & \textbf{24.743} & \textbf{-0.099} \\

  & Ours ($p=2.0$)
  & 0.0026 & 26.532 & 0.183 & \textbf{0.891} & 19.535 & -0.541
  & 0.0014 & 29.130 & 0.120 & 0.937 & 24.564 & -0.123 \\

  & Ours ($p=2.5$)
  & \textbf{0.0023} & \textbf{27.030} & \textbf{0.181} & 0.883 & 19.246 & -0.605
  & \textbf{0.0011} & \textbf{30.168} & \textbf{0.106} & \textbf{0.943} & 24.124
  & -0.131 \\

  \addlinespace[3pt]
  8 & Raw DDIM
  & 0.0070 & 21.954 & 0.259 & 0.819 & 19.119 & -0.627
  & 0.0065 & 22.255 & 0.191 & 0.874 & 23.508 & -0.231 \\

  & Ours ($p=1.7$)
  & 0.0041 & 24.506 & 0.186 & 0.886 & \textbf{20.235} & \textbf{-0.416}
  & 0.0036 & 24.920 & 0.144 & 0.902 & \textbf{24.856} & \textbf{-0.131} \\

  & Ours ($p=2.0$)
  & 0.0020 & 28.011 & 0.142 & \textbf{0.915} & 20.021 & -0.462
  & 0.0012 & 30.150 & 0.098 & 0.950 & 24.679 & -0.140 \\

  & Ours ($p=2.5$)
  & \textbf{0.0016} & \textbf{28.585} & \textbf{0.137} & 0.912 & 19.731 & -0.493
  & \textbf{0.0007} & \textbf{31.893} & \textbf{0.077} & \textbf{0.961} & 24.232
  & -0.155 \\

  \bottomrule
  \end{tabular}}
  \end{table*}

\subsection{Controlled Comparison of Input and Trajectory Correction}
  \label{app:bridge-input-trajectory}
\paragraph{Objective and comparison.}
  We investigate whether correcting the denoising updates can improve
  instance-wise fidelity beyond correcting only the initial noise.
  We compare raw three-step DDIM, input correction, and trajectory
  correction on CIFAR-10 and SD1.5. Within each model, the learned
  methods use the same pretrained backbone, paired data, reference
  sampler, three-step grid, endpoint objective, minibatch order for
  each training seed, and optimization-update budget.

  \paragraph{Correction parameterizations.}
  Input correction uses a second copy of the pretrained backbone with
  frozen base weights, rank-8 adapters, and a trainable zero-initialized
  output head. It predicts a residual adjustment to the initial noise;
  the corrected noise is normalized to the original per-sample norm
  before entering the unchanged three-step sampler. Trajectory correction
  uses a separate rank-8 adapter set at each of the three denoising
  steps, keeping the original initial noise and pretrained base weights
  fixed. Adapters target attention and convolutional projections on
  CIFAR-10 and attention projections on SD1.5. Both corrections are
  initialized to recover raw sampling and trained through the complete
  student rollout without intermediate-state supervision.

  \paragraph{Data and sampling.}
  CIFAR-10 uses a 1,000-step DDIM reference and student timesteps
  $[666,333,0]$. We use 4,096 training pairs, 256 validation pairs,
  and 1,000 held-out test noises, with 4,096 updates at batch size 32.
  SD1.5 uses a 100-step DDIM reference at CFG~1 and the native trailing
  three-step grid $[999,666,332]$. Training uses 2,048 pairs and
  2,048 updates at batch size four; validation uses 128 pairs.
  Testing uses 64 held-out prompts with four new noises each.

  \paragraph{Training and checkpoint selection.}
  Both methods minimize RGB MSE on CIFAR-10. On SD1.5, both minimize
  \begin{equation}
    \mathcal{L}
    = \frac{1}{B}\sum_{i=1}^{B}
      \left[0.1\log(m_{x,i}+10^{-6})+0.05m_{h,i}\right],
  \end{equation}
  where $m_{x,i}$ is continuous RGB MSE on the nominal $[-1,1]$
  scale and $m_{h,i}$ is terminal-latent MSE. The pretrained VAE
  weights remain frozen while gradients through its input are retained.
  This auxiliary term requires the reference terminal latent.

  All runs use AdamW with weight decay $10^{-4}$, gradient clipping
  at norm 5, and EMA decay 0.99. A 64-update validation screen between
  learning rates $10^{-4}$ and $3\times10^{-4}$ selects
  $3\times10^{-4}$ for both methods on both models. 
  We train two seeds per learned configuration. Validation PSNR
  selects among current and EMA weights evaluated every 256 updates;
  test data is not used for checkpoint or hyperparameter selection.

  \paragraph{Results.}
  Table~\ref{tab:bridge-input-trajectory} reports mean per-image
  metrics, averaged over the two training seeds for learned methods.
  MSE and PSNR use continuous RGB without clipping or quantization;
  reported MSE is normalized to the $[0,1]$ scale. LPIPS uses AlexNet
  at native resolution with RGB clipped to $[-1,1]$. SSIM uses clipped
  $[0,1]$ RGB with Gaussian weighting.

  \begin{table}[t]
    \centering
    \small
    \setlength{\tabcolsep}{4pt}
    \caption{Controlled comparison under matched data, endpoint
    supervision, and training-update budgets. MSE is multiplied by
    $10^3$; PSNR is in dB. Params counts trainable parameters, and Calls
    counts U-Net evaluations. The input arm uses one additional full
    U-Net corrector. References use 1,000 steps for CIFAR-10 and
    100 steps for SD1.5. Results use 1,000 CIFAR-10 test inputs or
    64 SD1.5 prompts with four noises each.}
    \label{tab:bridge-input-trajectory}
    \vspace{\baselineskip}
    \resizebox{\linewidth}{!}{\begin{tabular}{lcrrrrr}
      \toprule
      Method & Calls & Params (M) & MSE $\downarrow$ &
      PSNR $\uparrow$ & LPIPS $\downarrow$ & SSIM $\uparrow$ \\
      \midrule
      \multicolumn{7}{c}{CIFAR-10} \\
      \midrule
      Raw DDIM3
      & 3 & 0 & 21.18 & 17.29 & 0.149 & 0.578 \\
      Input correction
      & $3+1$ & 1.266 & 2.86 & 26.60 & 0.025 & 0.901 \\
      Trajectory correction
      & 3 & 3.788 & \textbf{1.66} & \textbf{29.79}
      & \textbf{0.011} & \textbf{0.944} \\
      \midrule
      \multicolumn{7}{c}{SD1.5, CFG 1} \\
      \midrule
      Raw DDIM3
      & 3 & 0 & 20.05 & 17.33 & 0.801 & 0.402 \\
      Input correction
      & $3+1$ & 1.606 & 9.70 & 20.60 & 0.408 & 0.570 \\
      Trajectory correction
      & 3 & 4.783 & \textbf{5.57} & \textbf{23.25}
      & \textbf{0.189} & \textbf{0.770} \\
      \bottomrule
    \end{tabular}}
  \end{table}

  Both correction methods improve all four fidelity metrics over raw
  sampling. Trajectory correction further improves PSNR over input
  correction by 3.19\,dB on CIFAR-10 and 2.65\,dB on SD1.5, with
  paired 95\% bootstrap intervals of $[3.03,3.35]$ and $[2.47,2.84]$.
  It also reduces LPIPS from 0.025 to 0.011 and from 0.408 to 0.189,
  respectively. Bootstrap intervals use 5,000 paired resamples of
  CIFAR inputs or SD prompts after averaging the two trained models;
  they quantify test-input uncertainty conditional on those models.
  The results demonstrate effective correction at both locations,
  with trajectory correction achieving higher fidelity under this
  training protocol.

  \paragraph{Computational cost and interpretation.}
  The parameterizations differ in trainable parameter count and
  implementation cost. Input correction adds one full U-Net call;
  trajectory correction operates within the existing three calls.

\subsection{Cross-Backbone Transfer}
\label{app:bedroom_transfer}
We transfer a frozen noise corrector trained with a Bedroom diffusion
  backbone to CelebA-HQ DDIM3, without retraining or test-time scale selection. All methods are evaluated on the same 200 noise seeds at $256\times256$
  resolution, which are disjoint from Table ~\ref{tab:flexible_steps}.
  The corrector was trained using cached CelebA 20-step DDIM teacher targets, so this
  experiment evaluates cross-backbone transfer with target-domain supervision.

\begin{table}[htbp!]
    \centering
    \caption{
    Cross-backbone transfer from Bedroom to CelebA-HQ.
    Results are mean $\pm$ SD over 200 samples.
    NFEs denote denoiser plus corrector forward passes.
    HPS denotes HPSv2.1 $\times100$; IR denotes ImageReward.
    Preference scores use the fixed text ``a portrait photo of a person''
    for all unconditional outputs.
    }
    \label{tab:bedroom_celeba_transfer}
    \resizebox{\linewidth}{!}{\begin{tabular}{lrrrr}
    \toprule
    \multicolumn{5}{l}{\textit{(a) Preference scores}} \\
    Method & NFEs & PickScore $\uparrow$ & HPS $\uparrow$ & IR $\uparrow$ \\
    \midrule
    Raw DDIM3
    & $3$
    & $19.03 \pm 0.45$
    & $16.53 \pm 2.22$
    & $-0.98 \pm 0.69$ \\
    Bedroom-backbone corrector  + DDIM3
    & $3+1$
    & $\mathbf{20.37} \pm 0.54$
    & $\mathbf{27.34} \pm 1.77$
    & $\mathbf{0.54} \pm 0.36$ \\
    CelebA corrector  + DDIM3
    & $3+1$
    & $19.90 \pm 0.62$
    & $24.04 \pm 2.13$
    & $0.10 \pm 0.50$ \\
    DDIM20 teacher
    & $20$
    & $19.89 \pm 0.61$
    & $24.88 \pm 2.13$
    & $0.10 \pm 0.51$ \\
    \midrule
    \multicolumn{5}{l}{\textit{(b) Agreement with the DDIM20 teacher}} \\
    Method & MSE ($\times10^{-3}$) $\downarrow$
    & PSNR $\uparrow$ & LPIPS $\downarrow$ & SSIM $\uparrow$ \\
    \midrule
    Raw DDIM3
    & $36.26 \pm 14.81$
    & $14.78 \pm 1.86$
    & $0.58 \pm 0.06$
    & $0.65 \pm 0.08$ \\
    Bedroom-backbone corrector + DDIM3
    & $5.43 \pm 1.53$
    & $22.82 \pm 1.20$
    & $0.27 \pm 0.06$
    & $0.81 \pm 0.04$ \\
    CelebA corrector  + DDIM3
    & $\mathbf{1.32} \pm 0.70$
    & $\mathbf{29.35} \pm 2.18$
    & $\mathbf{0.17} \pm 0.05$
    & $\mathbf{0.92} \pm 0.03$ \\
    \bottomrule
    \end{tabular}}
    \end{table}

  As shown in Table~\ref{tab:bedroom_celeba_transfer}, the transferred corrector improves PickScore, HPS, and ImageReward over raw DDIM3 by approximately 1.34, 10.81, and 1.52, respectively, using three denoiser evaluations and one corrector evaluation.
  In particular, it achieves higher mean preference scores than the evaluated
  CelebA corrector and DDIM20 teacher.
  Agreement with the teacher improves over raw DDIM3 by 8.04 dB in PSNR, with an approximately 53.4\% reduction in LPIPS.
  The CelebA corrector trained in Section ~\ref{subsec:noise corr} achieves closer teacher agreement across all four
  fidelity metrics. Therefore, with the same target, the in-domain corrector achieves much higher teacher fidelity,
  while transferring a corrector trained with an out-of-domain backbone achieves much
  higher aesthetic preference scores.

\subsection{Additional FLUX Evaluations}
\label{app:flux-additional}

\begin{figure}[htbp!]
  \centering
  \includegraphics[page=1,width=\linewidth]{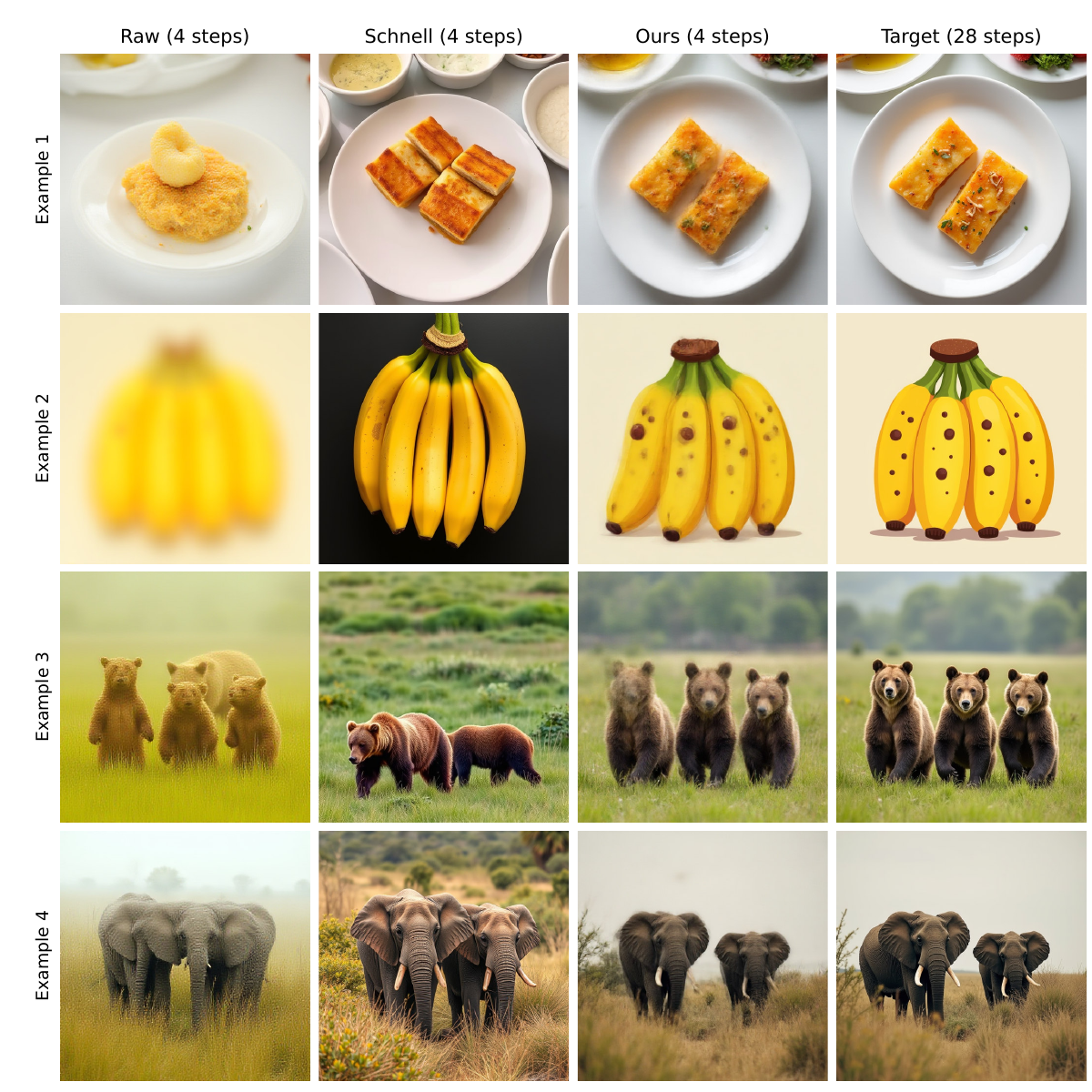}
  \caption{ Comparison of FLUX.1-schnell and our method as previews of FLUX.1-dev.
  Columns show raw four-step Euler, Schnell, our corrected four-step
  sampler, and the 28-step reference. Although Schnell achieves higher
  ImageReward on the previews themselves, our method provides greater
  fidelity to the reference and better candidate-selection performance.}
  \label{fig:app-schnell}
\end{figure}

\paragraph{Comparison with FLUX.1-schnell.}
\label{app:flux-schnell}
 We compare our method with a distilled few-step generator to investigate whether distillation provides an effective alternative for faithful previews. We evaluate four-step FLUX.1-schnell on
  256 prompts with eight seeds each, using the same prompts and
  initial noises as the FLUX.1-dev reference.
  Although Schnell achieves higher ImageReward on the previews
  themselves, our method produces substantially closer matches to
  the 28-step reference, achieving 18.954\,dB PSNR versus
  11.357\,dB. This advantage also improves candidate selection:
  selecting seeds by preview ImageReward yields final reference-image
  scores of 1.1361 with our method versus 0.9514 with Schnell
  (Figure~\ref{fig:app-schnell}).
  These results show that our method outperforms the evaluated
  distilled baseline in both preview fidelity and downstream  candidate selection.

\paragraph{Budget-allocation ablation.}
\label{app:flux-budget}
We conduct a small budget-allocation ablation to investigate how
  a fixed computation budget can be divided between candidate count
  and preview steps. On the same 256 prompts, we compare eight
  corrected four-step previews with four raw eight-step previews.
  Including the final 28-step run, both workflows require 60 NFE.
  Our method achieves a selected-image ImageReward of 1.1361 versus
  1.0573, a paired gain of 0.0788 (95\% CI $[0.0292,0.1308]$).
  These results show that screening more candidates with our
  corrected previews improves selection over spending more
  denoising steps on fewer uncorrected previews at the same budget.

\paragraph{Transfer across resolutions.}
\label{app:flux-resolution}
We study whether the learned correction generalizes to
  higher-resolution generation without retraining.
  We apply the frozen corrector trained at $512\times512$ directly
  at $768\times768$, retaining its four-step time grid.
  On 64 prompts with eight seeds each, our method achieves
  18.45\,dB PSNR, compared with 13.96\,dB for a raw control
  with the same guidance and time grid and 15.35\,dB for native
  eight-step Euler. Selected-image ImageReward is 1.247, compared
  with 1.125 and 1.154, respectively.
  These results show that the correction remains effective at a
  higher resolution, improving both reference fidelity  and
  ImageReward-based candidate selection without additional training.

\paragraph{Transfer to PartiPrompts.}
\label{app:flux-prompts}
We investigate whether the learned correction generalizes to
  a different prompt collection without retraining.
  We evaluate the same frozen checkpoint on 128 PartiPrompts with
  eight seeds each at $512\times512$. Our previews achieve
  19.082\,dB PSNR and 0.380 LPIPS, compared with 14.052\,dB and
  0.542 for native four-step Euler. ImageReward-based selection
  improves the final-image score from 1.3232 to 1.4102 and
  Spearman rank correlation from 0.079 to 0.357.
  These results show that the correction remains effective beyond
  the training prompt source, improving both reference fidelity
  and candidate selection on PartiPrompts.

\subsection{Ablation Studies}\label{app:subsec:ablation}
We evaluate frozen input correctors on 200 fixed noise samples per dataset, using 20-step DDIM outputs as references.

\begin{figure}[htbp!]
  \centering
  \includegraphics[width=\linewidth]{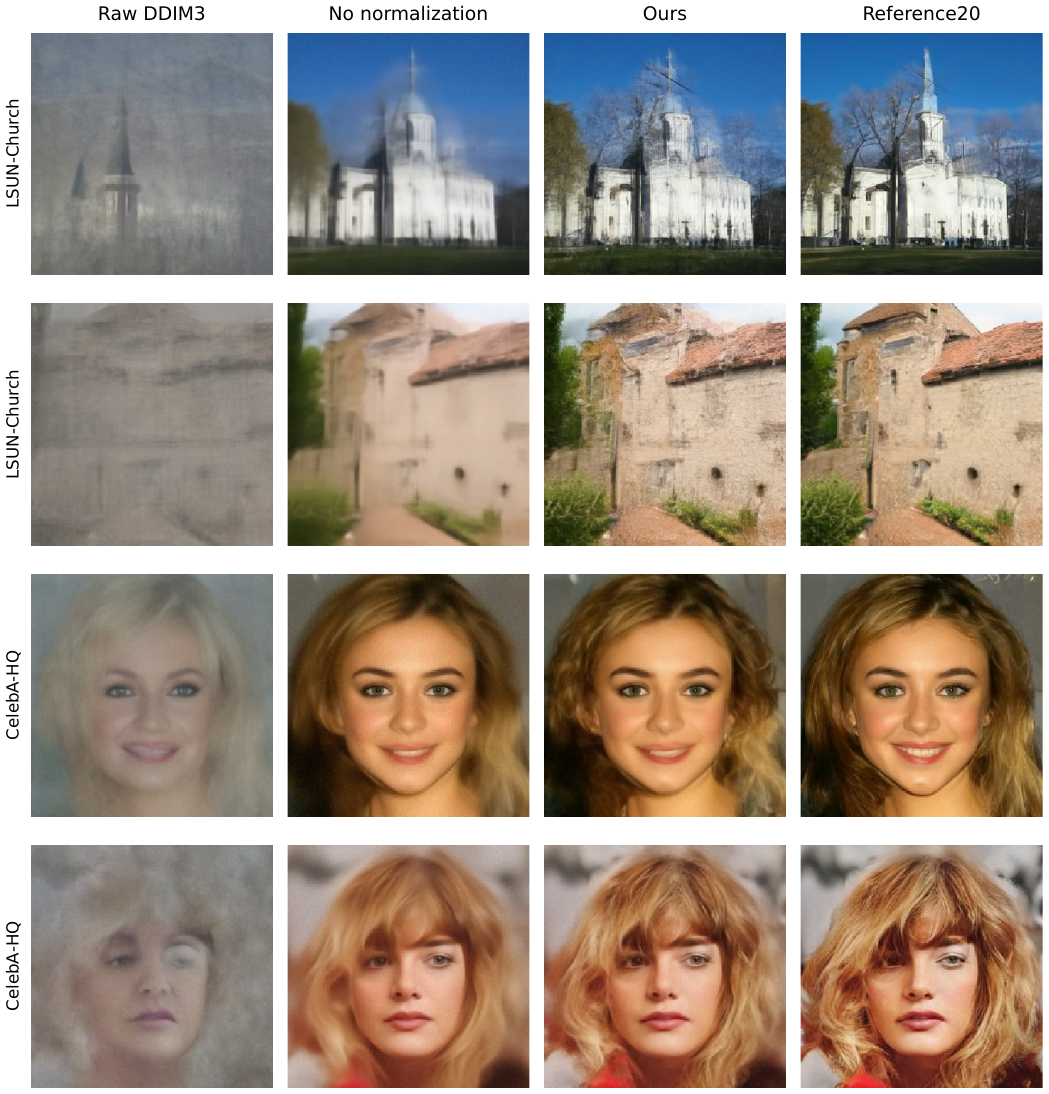}
  \caption{Noise norm normalization ablation. Columns show raw three-step DDIM, correction without normalization, complete correction, and the 20-step reference.}
  \label{fig:app-normalization}
\end{figure}

  \paragraph{Noise norm normalization.}
  We investigate whether preserving the original noise norm contributes to reconstruction fidelity. Removing normalization reduces three-step PSNR from 26.58 to 23.59\,dB on LSUN-Church and from 29.35 to 26.64\,dB on CelebA-HQ. Visually, removing norm normalization produces blurrier outputs
  with less distinct fine details (Figure~\ref{fig:app-normalization}). These results support normalizing the corrected noise to retain the original input norm. 

\begin{figure}[htbp!]
  \centering
  \includegraphics[width=\linewidth]{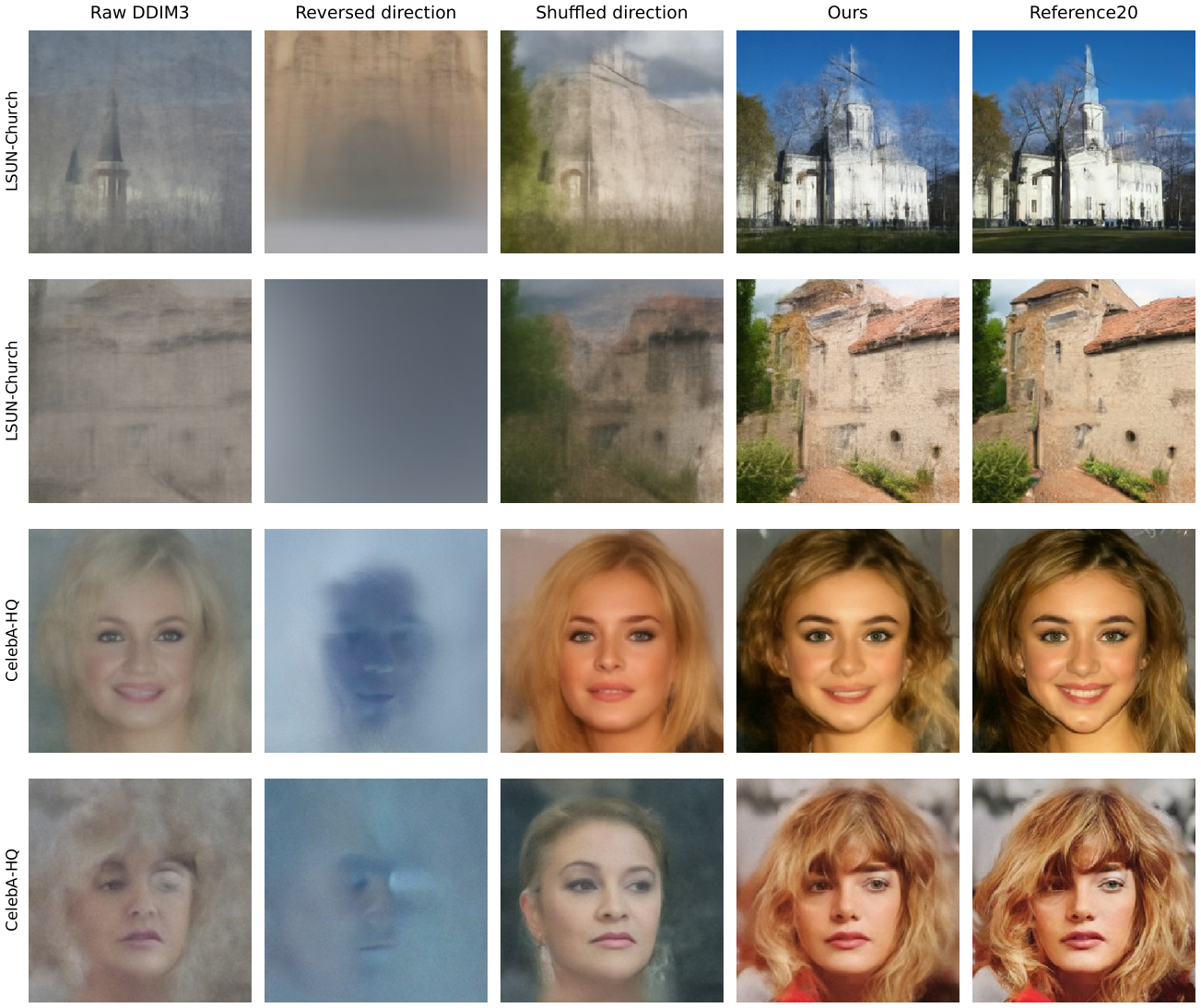}
  \caption{Correction direction ablation. Columns show raw three-step DDIM, reversed correction, another sample\textquotesingle{}s correction direction rescaled to the original residual norm, complete correction, and the 20-step reference.}
  \label{fig:app-direction}
\end{figure}

\paragraph{Sample-specific correction direction.}
  We investigate whether the improvement depends on the correction
  direction predicted for each input. We reverse the residual or
  replace its direction with that of another sample, preserving
  the original residual magnitude and applying noise norm
  normalization. Both controls yield lower PSNR than raw
  three-step DDIM (Figure~\ref{fig:app-direction}).
  This indicates that the gains depend on the sample-specific
  correction direction.

\begin{figure}[htbp!]
  \centering
  \includegraphics[width=\linewidth]{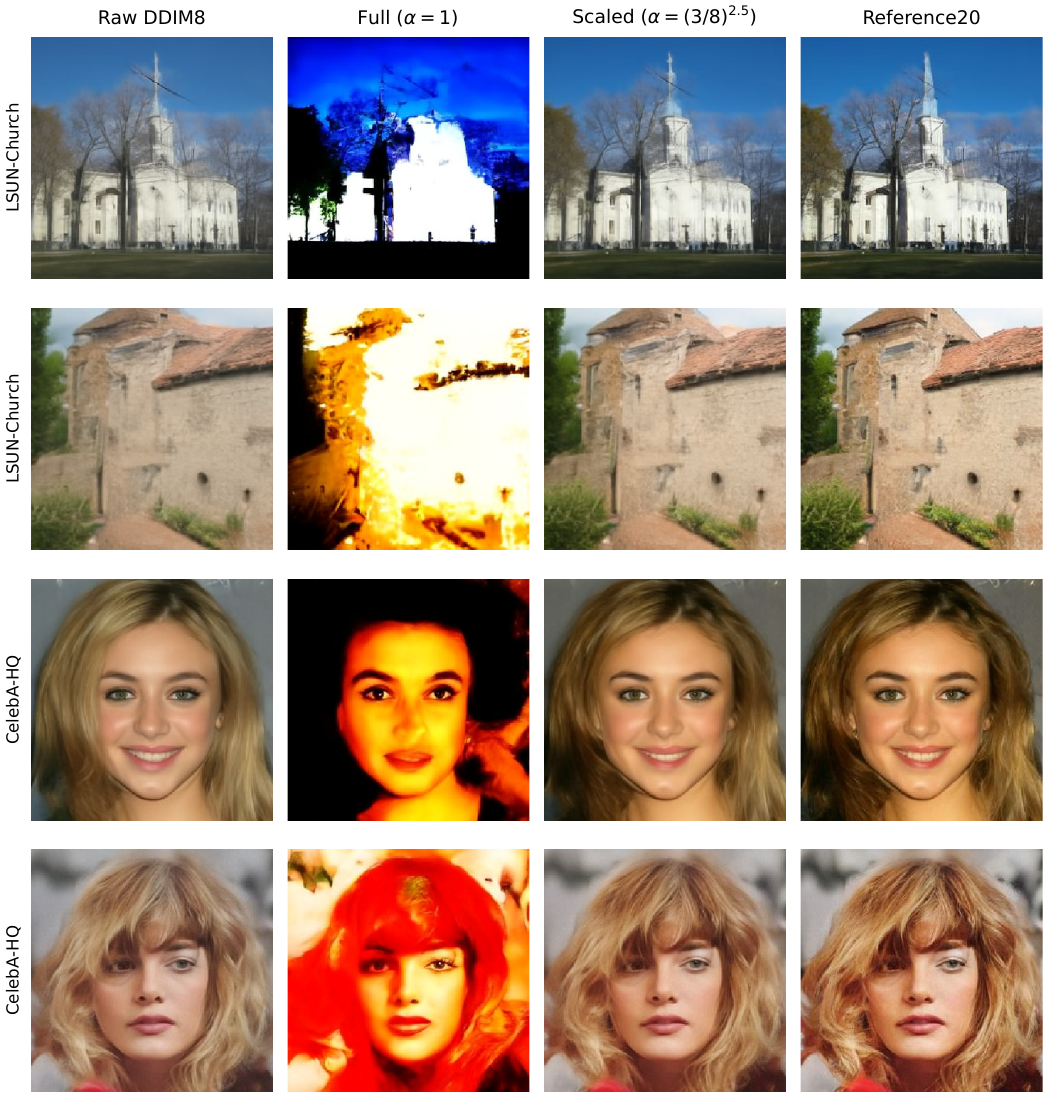}
  \caption{Attenuation when transferring a three-step corrector to eight-step DDIM. Columns show raw eight-step DDIM, correction with $\alpha=1$, correction with $\alpha=(3/8)^{2.5}$, and the 20-step reference.}
  \label{fig:app-attenuation}
\end{figure}

\paragraph{Attenuation across sampling steps.}
We test whether a input correction learned for three-step sampling
  should be reduced in magnitude when reused for eight-step sampling.
We apply the frozen
    three-step corrector to eight-step DDIM, comparing full-strength
    correction ($\alpha=1$) with attenuation
    ($\alpha=(3/8)^{2.5}$), while retaining norm normalization. We evaluate both settings on the same set of 200 noise samples
per dataset that are disjoint from previous experiments.
    Attenuation improves PSNR from 11.50 to 29.09\,dB on LSUN-Church
    and from 11.78 to 31.94\,dB on CelebA-HQ
    (Figure~\ref{fig:app-attenuation}).
    These results show that retaining the full correction can
    substantially degrade fidelity, supporting a smaller correction
    at larger sampling budgets.

\subsection{Runtime and Computational Cost}
 We measure inference time on a single NVIDIA GH200 GPU,
  averaging 32 measurements per setting after warm-up with CUDA synchronization.
  Three-step DDIM, our input-corrected sampler,
  and four-step DDIM take 87.6, 102.2, and 116.8\,ms on LSUN-Church,
  and 90.5, 109.4, and 121.4\,ms on CelebA-HQ, respectively.
  Thus, input correction costs less than an additional denoising
  step on these two datasets.

For FLUX.1-dev at $512\times512$ and batch size eight,
  four-step denoising takes 1.739\,s without trajectory correction
  and 1.740\,s with correction, indicating negligible inference
  overhead at the same NFE.
 
Each denoising step uses 104.60\,M LoRA parameters.
The four independent adapter sets contain 418.38\,M parameters in total, equivalent to 3.52\% of the frozen 11.90\,B-parameter FLUX transformer. 

\section{Additional Visualizations}
\label{app:visualizations}
\begin{figure}
    \centering
    \includegraphics[width=1\linewidth]{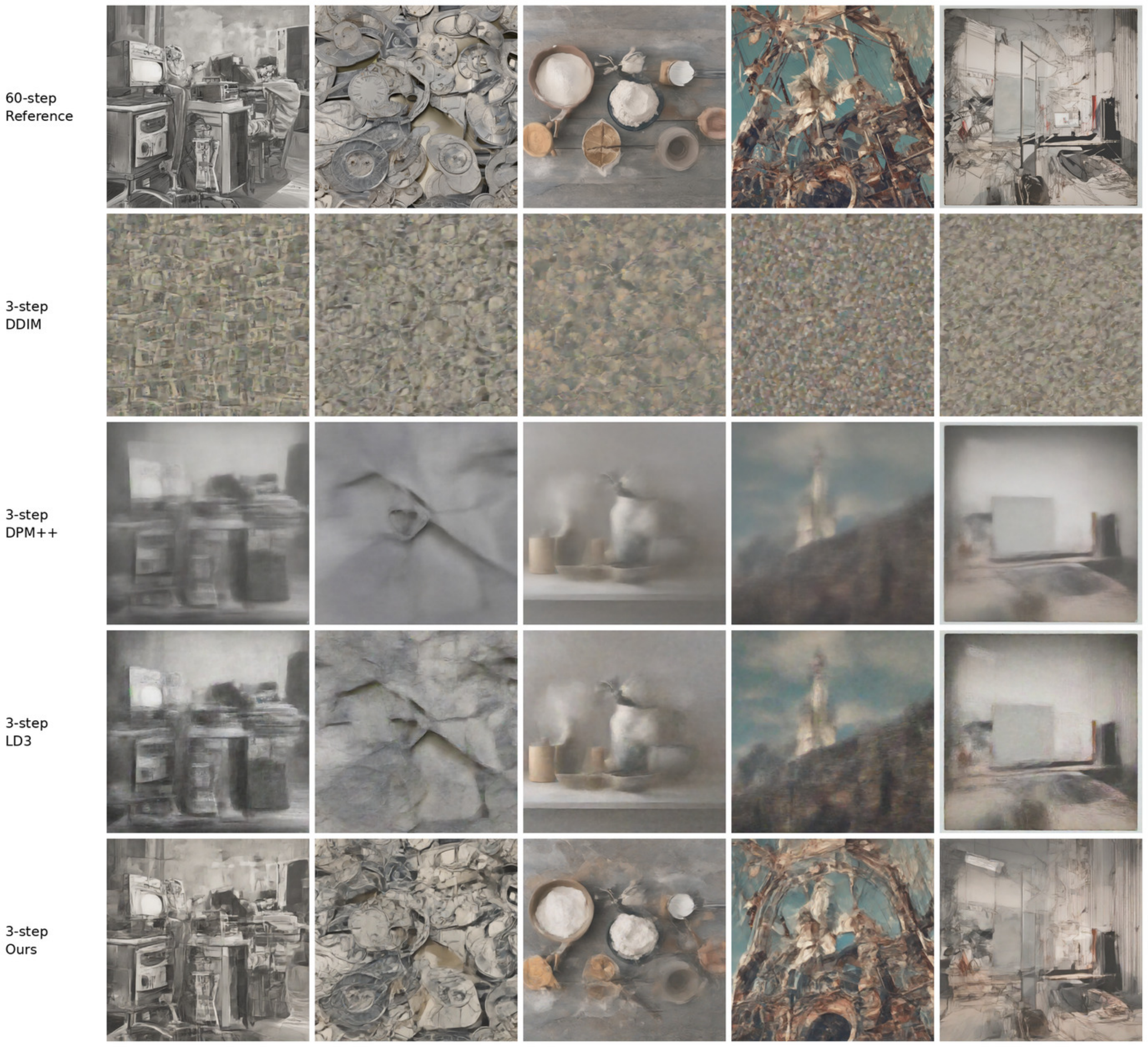}
    \caption{Visual comparison of 60-step Reference images and 3-step DDIM, DPM++, LD3, and Ours generations using SDXL at CFG 1.}
    \label{fig:sdxl_cfg1}
\end{figure}

\begin{figure}
    \centering
    \includegraphics[width=1\linewidth]{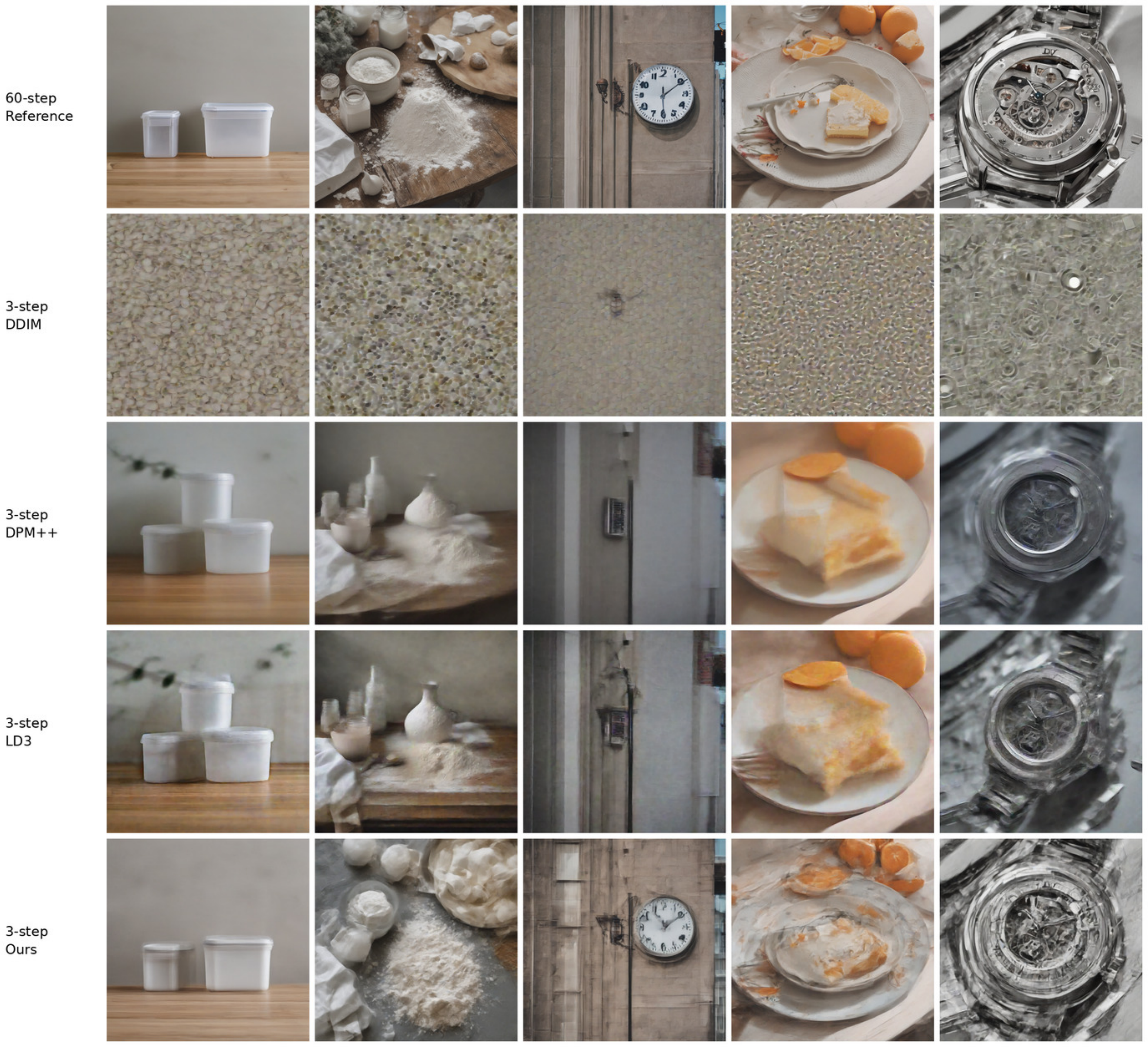}
    \caption{Visual comparison of 60-step Reference images and 3-step DDIM, DPM++, LD3, and Ours generations using SDXL at CFG 3.}
    \label{fig:sdxl_cfg3}
\end{figure}

\begin{figure}
    \centering
    \includegraphics[width=1\linewidth]{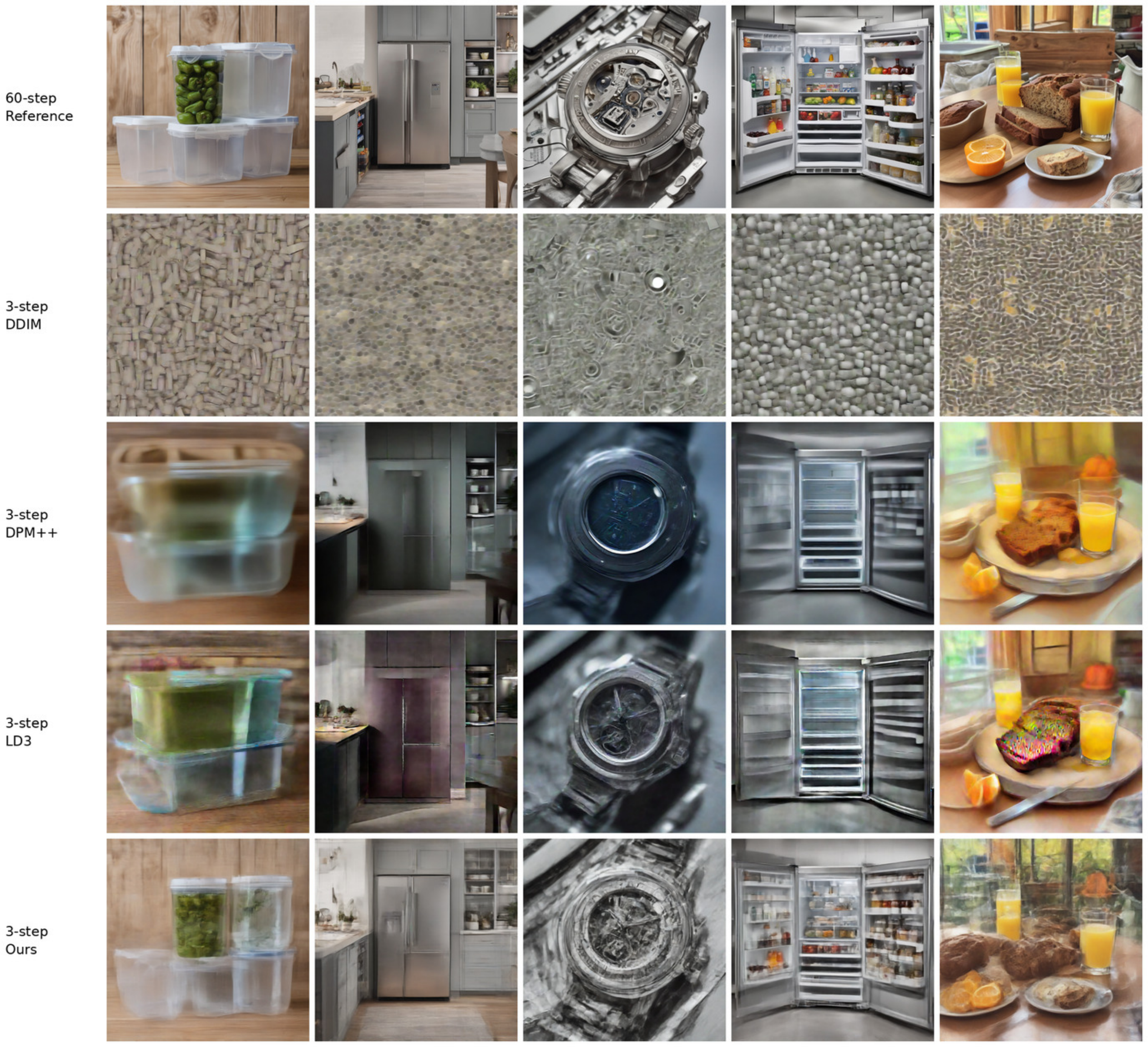}
    \caption{Visual comparison of 60-step Reference images and 3-step DDIM, DPM++, LD3, and Ours generations using SDXL at CFG 5.}
    \label{fig:sdxl_cfg5}
\end{figure}

\begin{figure}
    \centering
    \includegraphics[width=1\linewidth]{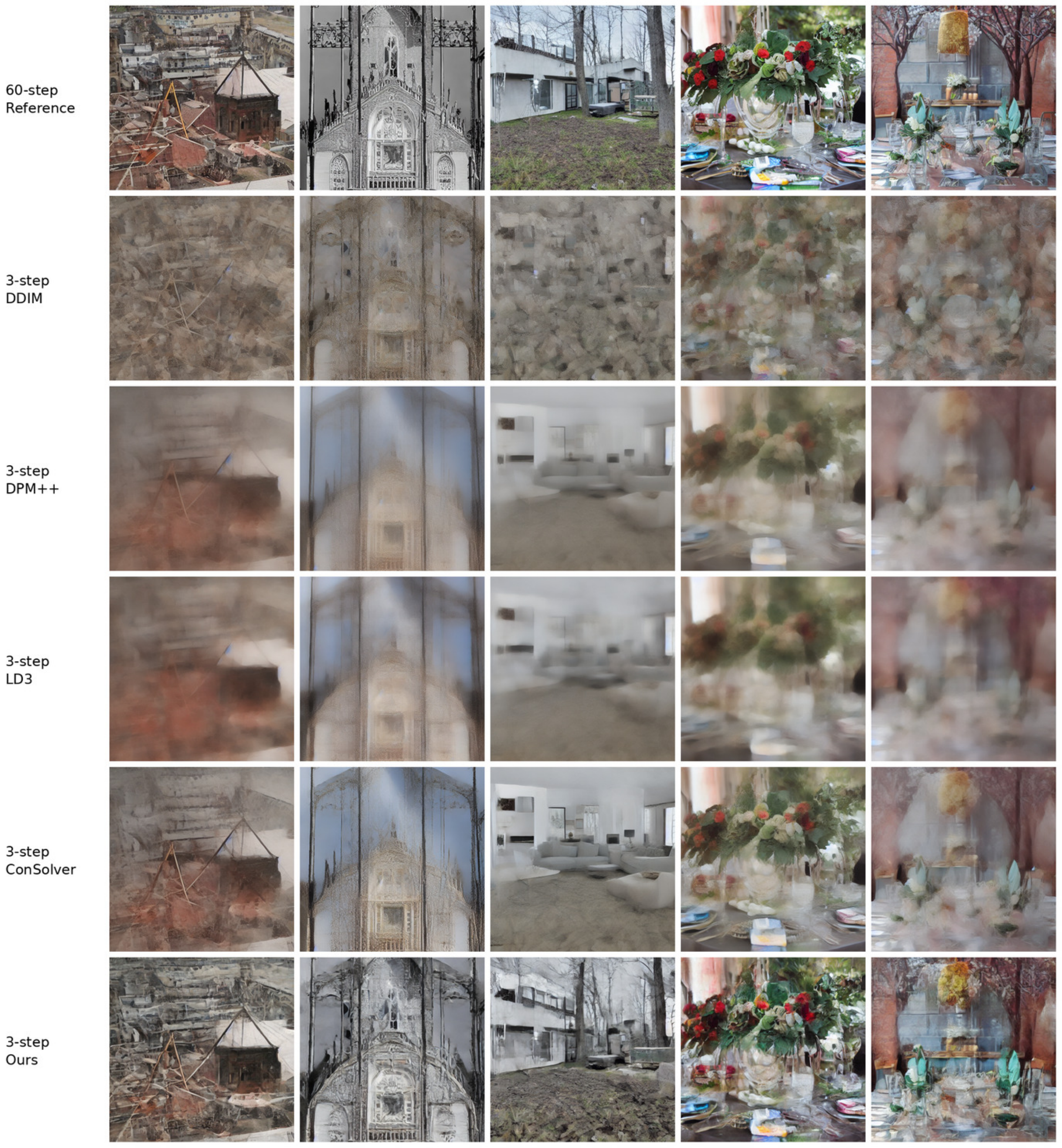}
    \caption{Visual comparison of 60-step Reference images and 3-step DDIM, DPM++, LD3, ConSolver, and Ours generations using Stable Diffusion 1.5 at CFG 1.}
    \label{fig:sd15_cfg1}
\end{figure}

\begin{figure}
    \centering
    \includegraphics[width=1\linewidth]{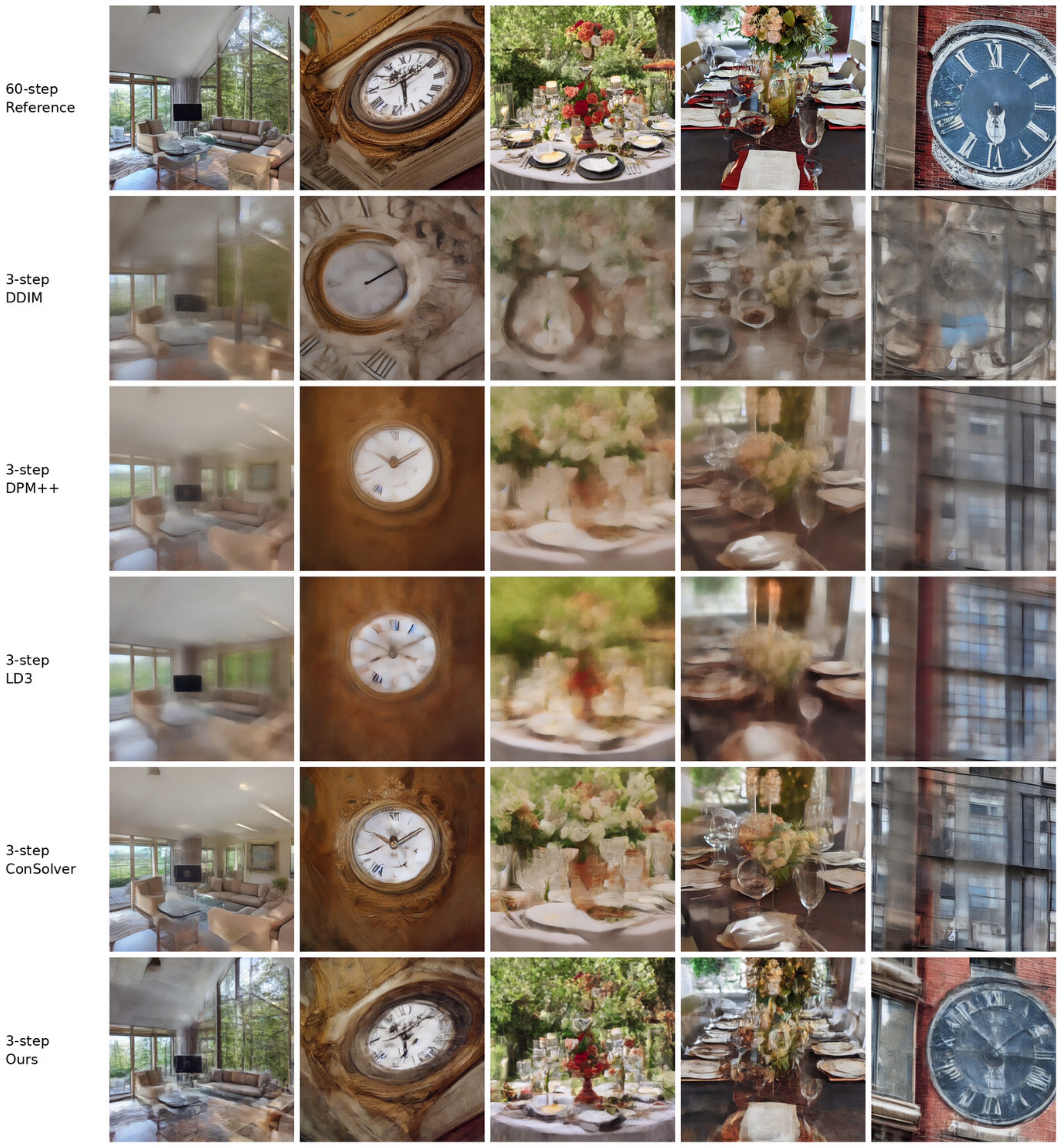}
    \caption{Visual comparison of 60-step Reference images and 3-step DDIM, DPM++, LD3, ConSolver, and Ours generations using Stable Diffusion 1.5 at CFG 3.}
    \label{fig:sd15_cfg3}
\end{figure}

\begin{figure}
    \centering
    \includegraphics[width=1\linewidth]{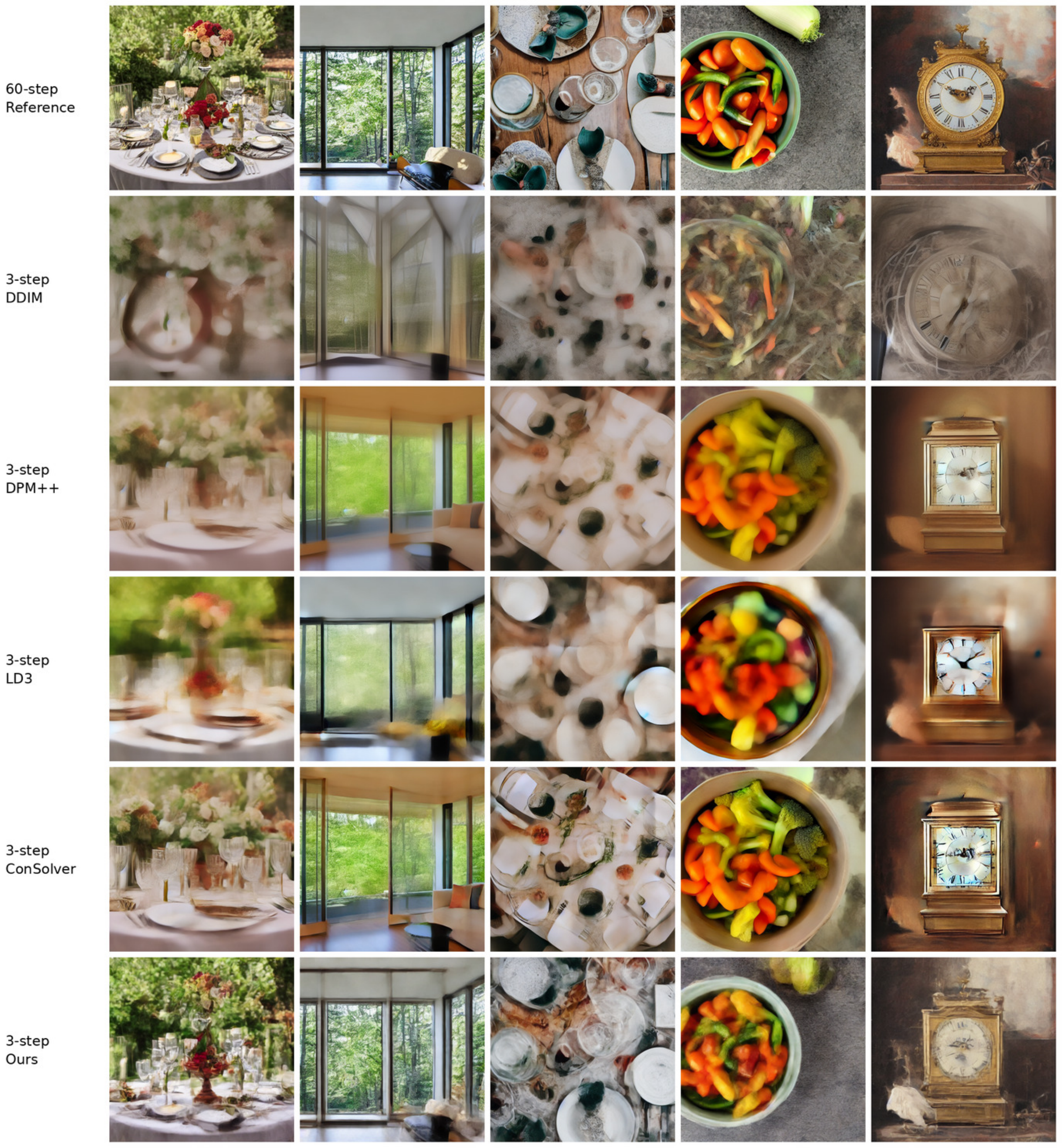}
    \caption{Visual comparison of 60-step Reference images and 3-step DDIM, DPM++, LD3, ConSolver, and Ours generations using Stable Diffusion 1.5 at CFG 5.}
    \label{fig:sd15_cfg5}
\end{figure}

\begin{figure}
    \centering
    \includegraphics[width=1\linewidth]{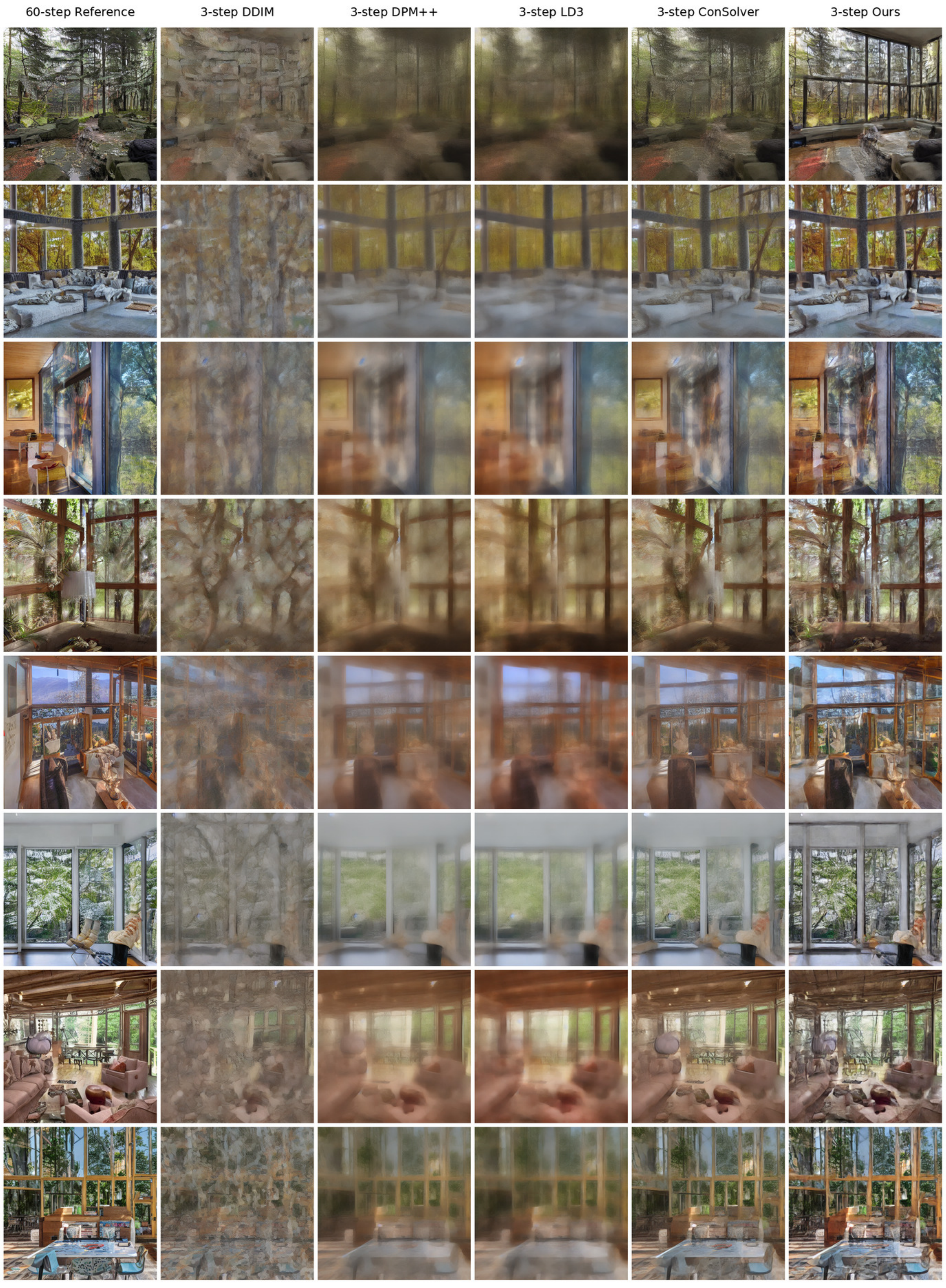}
    \caption{Comparison of 8 seed for the prompt “A large glass window in a living room.”}
    \label{fig:sd15_cfg1_group_000051}
\end{figure}

\begin{figure}
    \centering
    \includegraphics[width=0.8\linewidth]{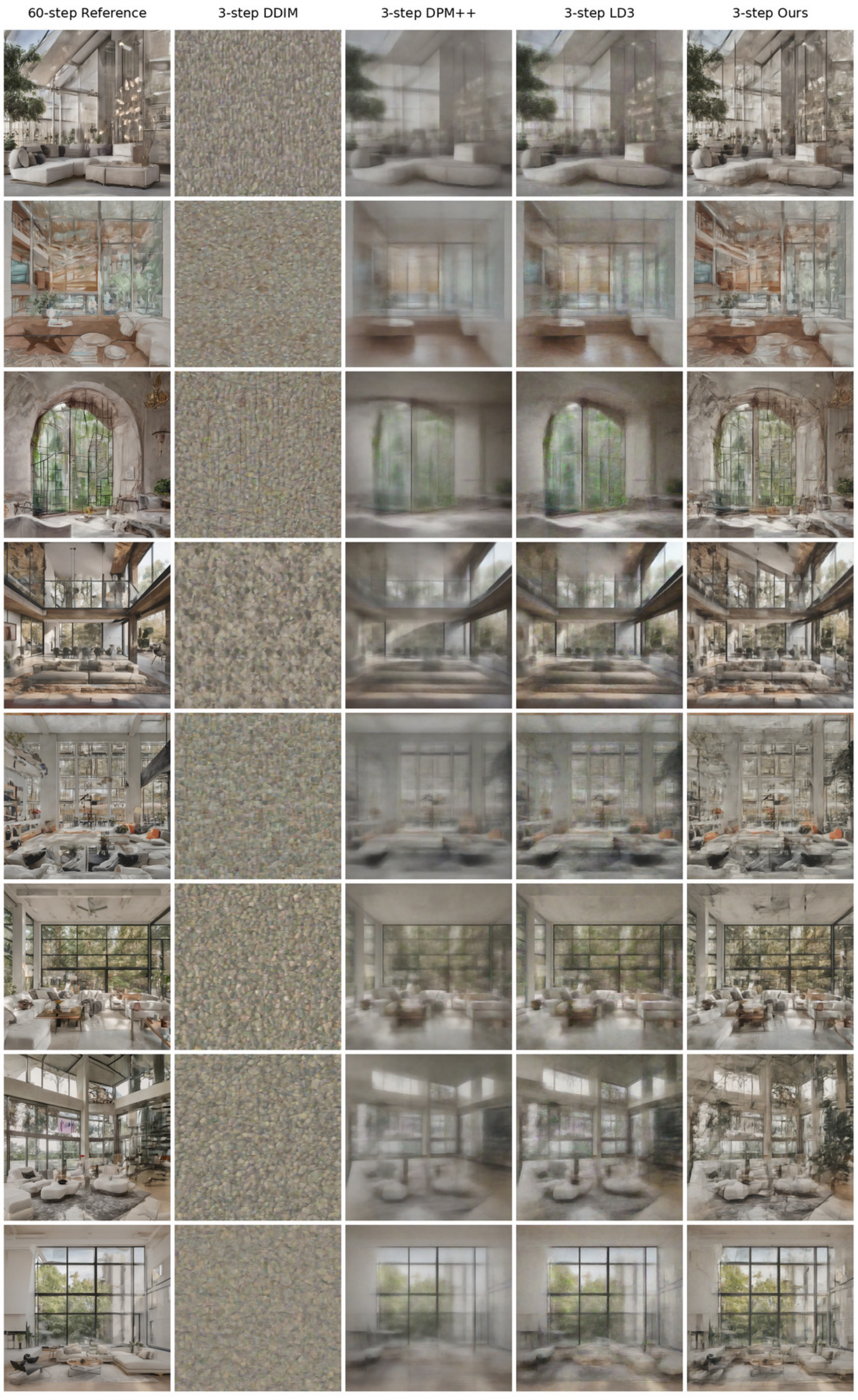}
    \caption{Comparison of 8 seed for the prompt “Comfortable, modern living room overlooking a wooded area”.}
    \label{fig:sdxl_cfg1_group_000039}
\end{figure}

\begin{figure}
    \centering
    \includegraphics[width=0.8\linewidth]{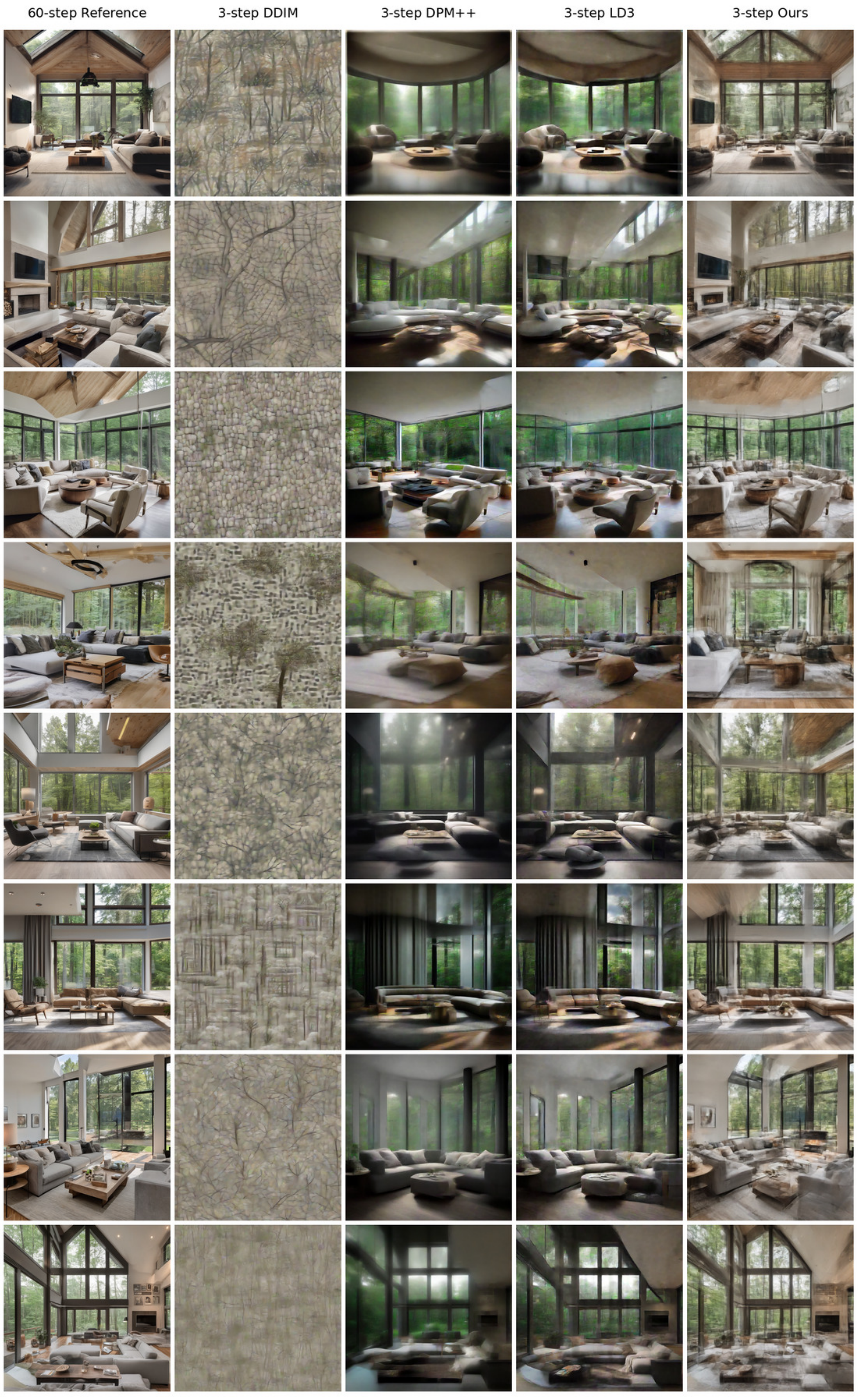}
    \caption{Comparison of 8 seed selection for the prompt “A living room with windows looking out onto a forest.”}
    \label{fig:sdxl_cfg5_group_000019}
\end{figure}

\subsection{Oracle Noise Optimization}
\begin{figure}[htbp!]
  \centering
  \includegraphics[width=\linewidth]{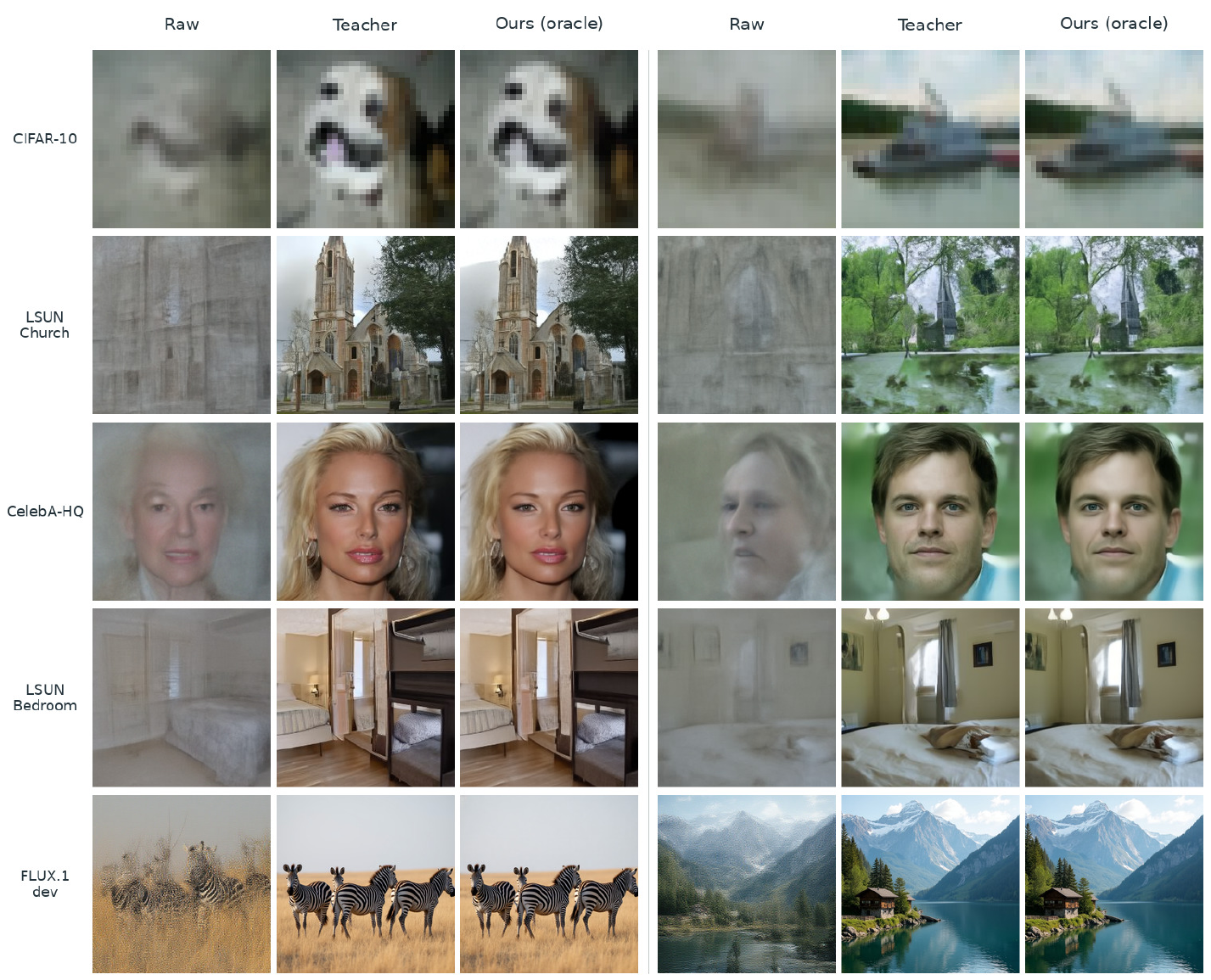}
  \caption{Additional oracle noise examples.}
  \label{fig:additional-oracle-noise-new-seeds}
\end{figure}

See Figure~\ref{fig:additional-oracle-noise-new-seeds}.

\subsection{Transfer Across Sampling Steps}
\begin{figure}[hbtp!]
  \centering
  \includegraphics[width=\linewidth]{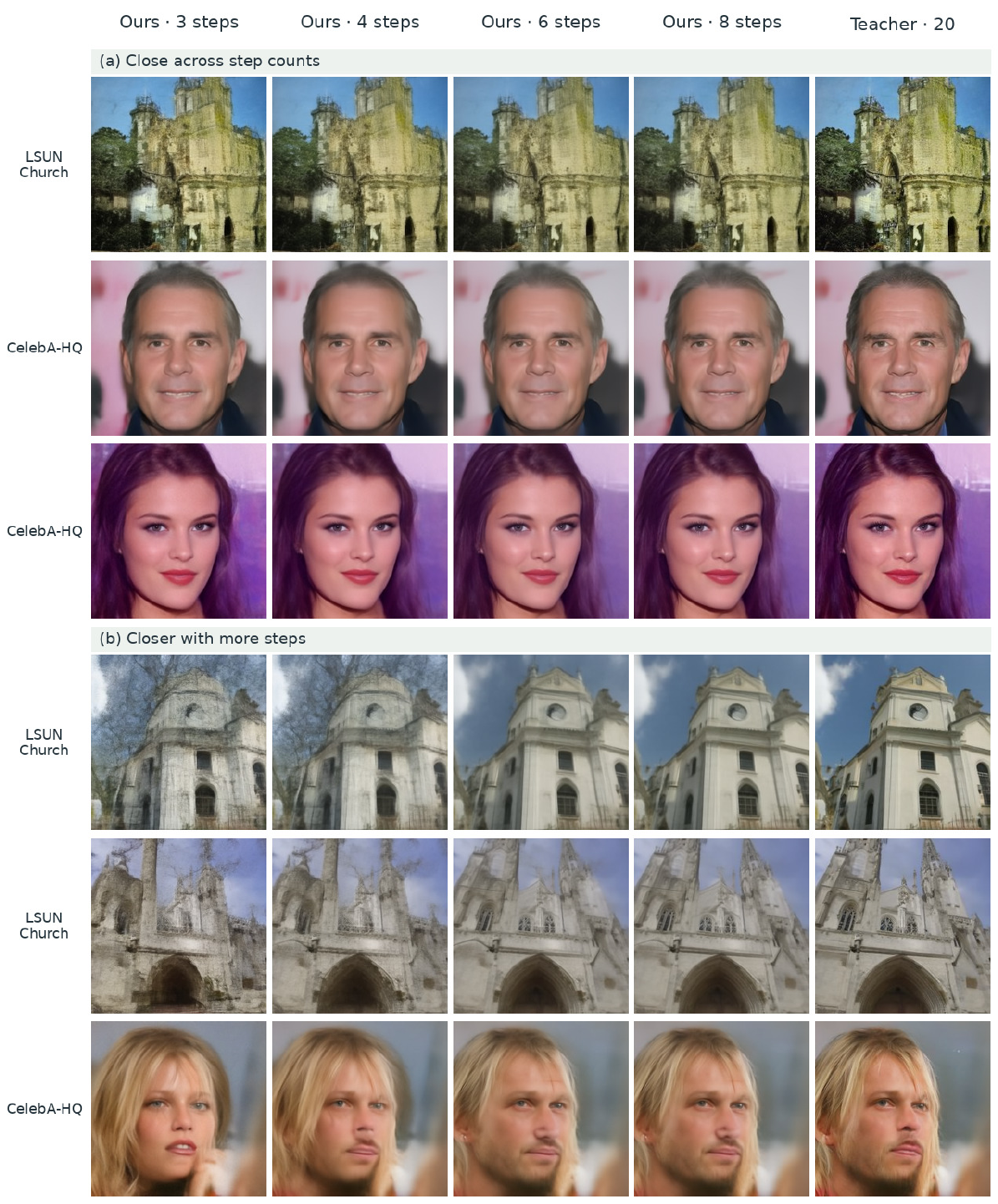}
  \caption{Selected examples of input-correction transfer across sampling steps. Top: examples remaining close to the teacher across step counts. Bottom: examples approaching the teacher as the step count increases. All outputs reuse the same frozen three-step corrector with a fixed exponent of 2.5 and noise norm normalization; each row shares the same original noise.}
  \label{fig:additional-step-transfer-new-seeds}
\end{figure}

See Figure~\ref{fig:additional-step-transfer-new-seeds}. 

\subsection{Cross-Backbone Transfer}
\begin{figure}[hbtp!]
  \centering
  \includegraphics[width=\linewidth]{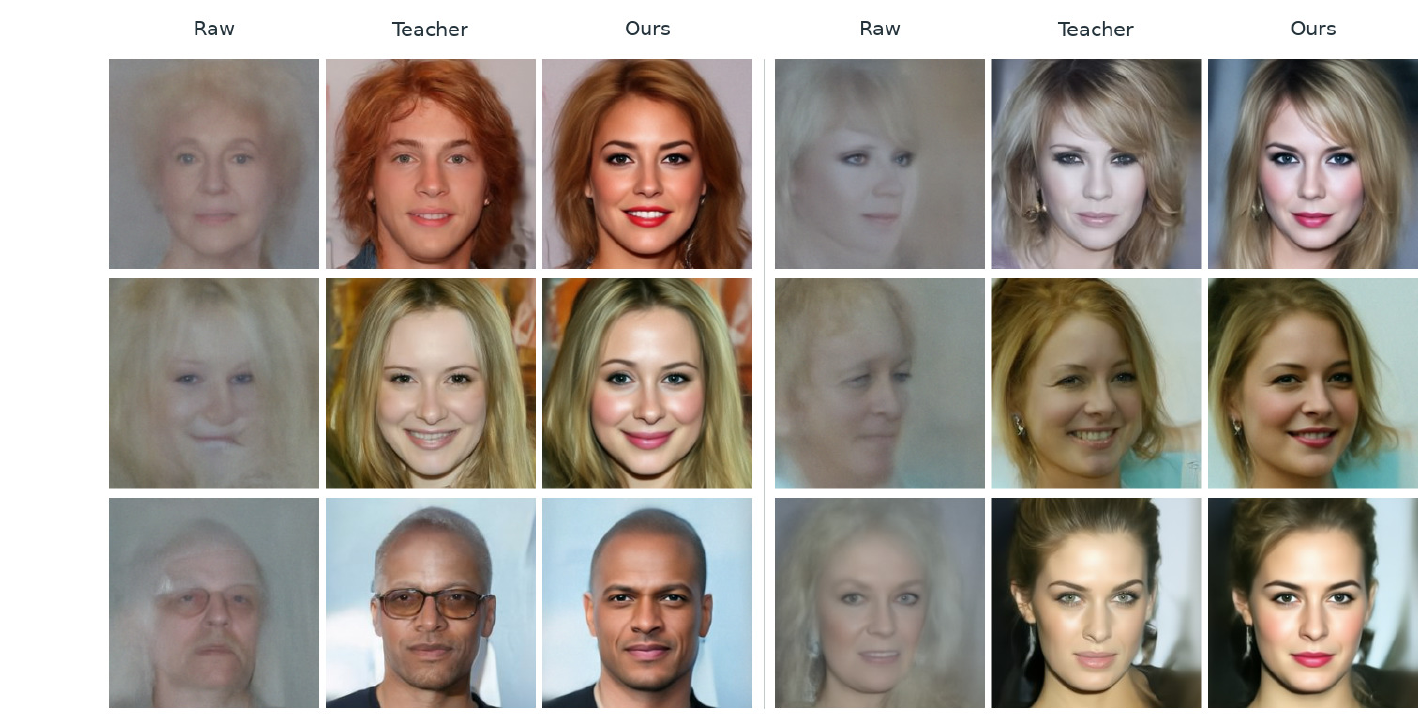}
  \caption{Additional cross-backbone correction examples on CelebA-HQ. Each triplet shows raw three-step DDIM, the paired 20-step teacher, and three-step DDIM with the transferred input corrector. The corrector was trained through a Bedroom backbone using CelebA teacher targets. }
  \label{fig:additional-cross-backbone-new-seeds}
\end{figure}

See Figure~\ref{fig:additional-cross-backbone-new-seeds}.

\end{document}

%% file: math_commands.tex
\usepackage{amsmath,amsfonts,bm}

\def\eqref#1{equation~\ref{#1}}

\def\1{\bm{1}}

\DeclareMathAlphabet{\mathsfit}{\encodingdefault}{\sfdefault}{m}{sl}
\SetMathAlphabet{\mathsfit}{bold}{\encodingdefault}{\sfdefault}{bx}{n}

%% file: references.bib
@inproceedings{podell2024sdxl,
  title={{SDXL}: Improving latent diffusion models for high-resolution image synthesis},
  author={Podell, Dustin and English, Zion and Lacey, Kyle and Blattmann, Andreas and Dockhorn, Tim and M{\"u}ller, Jonas and Penna, Joe and Rombach, Robin},
  booktitle={International Conference on Learning Representations},
  volume={2024},
  pages={1862--1874},
  year={2024}
}

@inproceedings{
	esser2024scaling,
	title={Scaling Rectified Flow Transformers for High-Resolution Image Synthesis},
	author={Patrick Esser and Sumith Kulal and Andreas Blattmann and Rahim Entezari and Jonas M{\"u}ller and Harry Saini and Yam Levi and Dominik Lorenz and Axel Sauer and Frederic Boesel and Dustin Podell and Tim Dockhorn and Zion English and Robin Rombach},
	booktitle={Forty-first International Conference on Machine Learning},
	year={2024},
	url={https://openreview.net/forum?id=FPnUhsQJ5B}
}

@article{cai2025z,
  title={Z-image: An efficient image generation foundation model with single-stream diffusion transformer},
  author={{Z-Image Team} and Huanqia Cai and Sihan Cao and Ruoyi Du and Peng Gao and Aiming Hao and Steven Hoi and Zhaohui Hou and Shijie Huang and Dengyang Jiang and Yuming Jiang and Xin Jin and Liangchen Li and Zhen Li and Zhong-Yu Li and David Liu and Dongyang Liu and Qilong Wu and Feng Yu and Zechao Zhan and Chi Zhang and Shifeng Zhang and Ruikai Zhou and Shilin Zhou},
  year={2026},
  eprint={2511.22699},
  archivePrefix={arXiv},
  primaryClass={cs.CV},
  url={https://arxiv.org/abs/2511.22699}, 
}

@article{labs2025flux,
  title={Flux. 1 kontext: Flow matching for in-context image generation and editing in latent space},
  author={{Black Forest Labs} and Stephen Batifol and Andreas Blattmann and Frederic Boesel and Saksham Consul and Cyril Diagne and Tim Dockhorn and Jack English and Zion English and Patrick Esser and Sumith Kulal and Kyle Lacey and Yam Levi and Cheng Li and Dominik Lorenz and Jonas Müller and Dustin Podell and Robin Rombach and Harry Saini and Axel Sauer and Luke Smith},
  year={2025},
  eprint={2506.15742},
  archivePrefix={arXiv},
  primaryClass={cs.GR},
  url={https://arxiv.org/abs/2506.15742}}

@article{lin2024sdxl,
  title={{SDXL}-lightning: Progressive adversarial diffusion distillation},
  author={Lin, Shanchuan and Wang, Anran and Yang, Xiao},
  journal={arXiv preprint arXiv:2402.13929},
  year={2024}
}

@inproceedings{yin2024one,
  title={One-step diffusion with distribution matching distillation},
  author={Yin, Tianwei and Gharbi, Micha{\"e}l and Zhang, Richard and Shechtman, Eli and Durand, Fredo and Freeman, William T and Park, Taesung},
  booktitle={2024 IEEE/CVF Conference on Computer Vision and Pattern Recognition (CVPR)},
  pages={6613--6623},
  year={2024},
  organization={IEEE}
}

@article{yin2024improved,
  title={Improved distribution matching distillation for fast image synthesis},
  author={Yin, Tianwei and Gharbi, Micha{\"e}l and Park, Taesung and Zhang, Richard and Shechtman, Eli and Durand, Fredo and Freeman, William T},
  journal={Advances in neural information processing systems},
  volume={37},
  pages={47455--47487},
  year={2024}
}

@article{lu2022dpm,
  title={{DPM}-solver: A fast ode solver for diffusion probabilistic model sampling in around 10 steps},
  author={Lu, Cheng and Zhou, Yuhao and Bao, Fan and Chen, Jianfei and Li, Chongxuan and Zhu, Jun},
  journal={Advances in neural information processing systems},
  volume={35},
  pages={5775--5787},
  year={2022}
}

@inproceedings{
salimans2022progressive,
title={Progressive Distillation for Fast Sampling of Diffusion Models},
author={Tim Salimans and Jonathan Ho},
booktitle={International Conference on Learning Representations},
year={2022},
url={https://openreview.net/forum?id=TIdIXIpzhoI}
}

@InProceedings{song2023consistency,
  title = 	 {Consistency Models},
  author =       {Song, Yang and Dhariwal, Prafulla and Chen, Mark and Sutskever, Ilya},
  booktitle = 	 {Proceedings of the 40th International Conference on Machine Learning},
  pages = 	 {32211--32252},
  year = 	 {2023},
  editor = 	 {Krause, Andreas and Brunskill, Emma and Cho, Kyunghyun and Engelhardt, Barbara and Sabato, Sivan and Scarlett, Jonathan},
  volume = 	 {202},
  series = 	 {Proceedings of Machine Learning Research},
  month = 	 {23--29 Jul},
  publisher =    {PMLR},
  url = 	 {https://proceedings.mlr.press/v202/song23a.html},
}

@inproceedings{wang2026image,
  title={Image diffusion preview with consistency solver},
    author    = {Wang, Fu-Yun and Zhou, Hao and Yuan, Liangzhe and Woo, Sanghyun and Gong, Boqing and Han, Bohyung and Yang, Ming-Hsuan and Zhang, Han and Zhu, Yukun and Liu, Ting and Zhao, Long},
  booktitle={Proceedings of the IEEE/CVF Conference on Computer Vision and Pattern Recognition},
  pages={43271--43280},
  year={2026}
}

@inproceedings{tong2025learning,
  title={Learning to discretize denoising diffusion odes},
  author={Tong, Vinh and Hoang, Trung-Dung and Liu, Anji and Van den Broeck, Guy and Niepert, Mathias},
  booktitle={International Conference on Learning Representations},
  volume={2025},
  pages={47244--47282},
  year={2025}
}

@InProceedings{huang2026diffusion,
    author    = {Huang, Bukun and Cui, Benlei and Ye, Zhizeng and Dong, Xuemei and Chen, Tuo and Xue, Hui and Yang, Dingkang and Huang, Longtao and Hong, Haiwen and Tang, Jingqun},
    title     = {Diffusion Probe: Generated Image Result Prediction Using CNN Probes},
    booktitle = {Proceedings of the IEEE/CVF Conference on Computer Vision and Pattern Recognition (CVPR)},
    month     = {June},
    year      = {2026},
    pages     = {35926-35935}
}

@InProceedings{guo2026toward,
    author    = {Guo, Huanlei and Wei, Hongxin and Jing, Bingyi},
    title     = {Toward Early Quality Assessment of Text-to-Image Diffusion Models},
    booktitle = {Proceedings of the IEEE/CVF Conference on Computer Vision and Pattern Recognition (CVPR)},
    month     = {June},
    year      = {2026},
    pages     = {38410-38419}
}

@article{eyring2024reno,
  title={Reno: Enhancing one-step text-to-image models through reward-based noise optimization},
  author={Eyring, Luca and Karthik, Shyamgopal and Roth, Karsten and Dosovitskiy, Alexey and Akata, Zeynep},
  journal={Advances in Neural Information Processing Systems},
  volume={37},
  pages={125487--125519},
  year={2024}
}

@inproceedings{
	ahn2026noise,
	title={A Noise is Worth Diffusion Guidance},
	author={Donghoon Ahn and Jiwon Kang and Sanghyun Lee and Jaewon Min and Minjae Kim and Wooseok Jang and Hyoungwon Cho and Sayak Paul and SeonHwa Kim and Eunju Cha and Kyong Hwan Jin and Seungryong Kim},
	booktitle={The Fourteenth International Conference on Learning Representations},
	year={2026},
	url={https://openreview.net/forum?id=xEWooSOgaz}
}

@inproceedings{
jia2026weak,
title={Weak Diffusion Priors Can Still Achieve Strong Inverse-Problem Performance},
author={Jing Jia and Wei Yuan and Sifan Liu and Liyue Shen and Guanyang Wang},
booktitle={Forty-third International Conference on Machine Learning},
year={2026},
url={https://openreview.net/forum?id=fdkSA4F0lN}
}

@article{wang2024dmplug,
  title={{DMP}lug: A plug-in method for solving inverse problems with diffusion models},
  author={Wang, Hengkang and Zhang, Xu and Li, Taihui and Wan, Yuxiang and Chen, Tiancong and Sun, Ju},
  journal={Advances in Neural Information Processing Systems},
  volume={37},
  pages={117881--117916},
  year={2024}
}

@InProceedings{Rombach_2022_CVPR,
    author    = {Rombach, Robin and Blattmann, Andreas and Lorenz, Dominik and Esser, Patrick and Ommer, Bj\"orn},
    title     = {High-Resolution Image Synthesis With Latent Diffusion Models},
    booktitle = {Proceedings of the IEEE/CVF Conference on Computer Vision and Pattern Recognition (CVPR)},
    month     = {June},
    year      = {2022},
    pages     = {10684-10695}
}

@inproceedings{hessel2021clipscore,
  title={Clipscore: A reference-free evaluation metric for image captioning},
  author={Hessel, Jack and Holtzman, Ari and Forbes, Maxwell and Le Bras, Ronan and Choi, Yejin},
  booktitle={Proceedings of the 2021 conference on empirical methods in natural language processing},
  pages={7514--7528},
  year={2021}
}

@inproceedings{
hu2022lora,
title={Lo{RA}: Low-Rank Adaptation of Large Language Models},
author={Edward J Hu and yelong shen and Phillip Wallis and Zeyuan Allen-Zhu and Yuanzhi Li and Shean Wang and Lu Wang and Weizhu Chen},
booktitle={International Conference on Learning Representations},
year={2022},
url={https://openreview.net/forum?id=nZeVKeeFYf9}
}

@inproceedings{lin2014microsoft,
  title={Microsoft coco: Common objects in context},
  author={Lin, Tsung-Yi and Maire, Michael and Belongie, Serge and Hays, James and Perona, Pietro and Ramanan, Deva and Doll{\'a}r, Piotr and Zitnick, C Lawrence},
  booktitle={European conference on computer vision},
  pages={740--755},
  year={2014},
  organization={Springer}
}

@article{wu2023human,
  title={Human preference score v2: A solid benchmark for evaluating human preferences of text-to-image synthesis},
  author={Wu, Xiaoshi and Hao, Yiming and Sun, Keqiang and Chen, Yixiong and Zhu, Feng and Zhao, Rui and Li, Hongsheng},
  journal={arXiv preprint arXiv:2306.09341},
  year={2023}
}

@article{yu2015lsun,
  title={{LSUN}: Construction of a large-scale image dataset using deep learning with humans in the loop},
  author={Yu, Fisher and Seff, Ari and Zhang, Yinda and Song, Shuran and Funkhouser, Thomas and Xiao, Jianxiong},
  journal={arXiv preprint arXiv:1506.03365},
  year={2015}
}

@inproceedings{
song2021denoising,
title={Denoising Diffusion Implicit Models},
author={Jiaming Song and Chenlin Meng and Stefano Ermon},
booktitle={International Conference on Learning Representations},
year={2021},
url={https://openreview.net/forum?id=St1giarCHLP}
}

@techreport{krizhevsky2009learning,
    title       = {Learning Multiple Layers of Features from Tiny Images},
    author      = {Krizhevsky, Alex},
    institution = {University of Toronto},
    year        = {2009},
    url         =
    {https://www.cs.toronto.edu/~kriz/learning-features-2009-TR.pdf}
  }

@inproceedings{liu2015faceattributes,
  title = {Deep Learning Face Attributes in the Wild},
  author = {Liu, Ziwei and Luo, Ping and Wang, Xiaogang and Tang, Xiaoou},
  booktitle = {Proceedings of International Conference on Computer Vision (ICCV)},
  month = {December},
  year = {2015} 
}

@inproceedings{NEURIPS2020_4c5bcfec,
 author = {Ho, Jonathan and Jain, Ajay and Abbeel, Pieter},
 booktitle = {Advances in Neural Information Processing Systems},
 editor = {H. Larochelle and M. Ranzato and R. Hadsell and M.F. Balcan and H. Lin},
 pages = {6840--6851},
 publisher = {Curran Associates, Inc.},
 title = {Denoising Diffusion Probabilistic Models},
 url = {https://proceedings.neurips.cc/paper_files/paper/2020/file/4c5bcfec8584af0d967f1ab10179ca4b-Paper.pdf},
 volume = {33},
 year = {2020}
}

@inproceedings{NEURIPS2023_73aacd8b,
	author = {Kirstain, Yuval and Polyak, Adam and Singer, Uriel and Matiana, Shahbuland and Penna, Joe and Levy, Omer},
	booktitle = {Advances in Neural Information Processing Systems},
	doi = {10.52202/075280-1594},
	editor = {A. Oh and T. Naumann and A. Globerson and K. Saenko and M. Hardt and S. Levine},
	pages = {36652--36663},
	publisher = {Curran Associates, Inc.},
	title = {Pick-a-Pic: An Open Dataset of User Preferences for Text-to-Image Generation},
	url = {https://proceedings.neurips.cc/paper_files/paper/2023/file/73aacd8b3b05b4b503d58310b523553c-Paper-Conference.pdf},
	volume = {36},
	year = {2023}
}

@article{lu2025dpmsolverpp,
	title={{DPM-Solver++}: Fast Solver for Guided Sampling of Diffusion Probabilistic Models},
	author={Lu, Cheng and Zhou, Yuhao and Bao, Fan and Chen, Jianfei and Li, Chongxuan and Zhu, Jun},
	journal={Machine Intelligence Research},
	volume={22},
	number={4},
	pages={730--751},
	year={2025},
	doi={10.1007/s11633-025-1562-4},
	url={https://doi.org/10.1007/s11633-025-1562-4}
}

@inproceedings{
	xu2023imagereward,
	title={ImageReward: Learning and Evaluating Human Preferences for Text-to-Image Generation},
	author={Jiazheng Xu and Xiao Liu and Yuchen Wu and Yuxuan Tong and Qinkai Li and Ming Ding and Jie Tang and Yuxiao Dong},
	booktitle={Thirty-seventh Conference on Neural Information Processing Systems},
	year={2023},
	url={https://openreview.net/forum?id=JVzeOYEx6d}
}
